\documentclass[runningheads]{llncs}
\usepackage[T1]{fontenc}
\usepackage{graphicx}
\usepackage{caption}
\usepackage{subcaption}
\usepackage{algorithm}
\usepackage[noend]{algpseudocode}
\usepackage{hyperref}
\usepackage[depth=2]{bookmark}
\usepackage{color}

\usepackage{amsfonts}
\usepackage{amsmath}
\usepackage{comment}

\usepackage{color,xcolor}
\definecolor{MyDarkBlue}{rgb}{0,0.08,1}
\definecolor{MyDarkGreen}{rgb}{0.02,0.6,0.02}
\definecolor{MyDarkRed}{rgb}{0.8,0.02,0.02}
\definecolor{MyDarkOrange}{rgb}{0.40,0.2,0.02}
\definecolor{MyPurple}{RGB}{111,0,255}
\definecolor{MyRed}{rgb}{1.0,0.0,0.0}
\definecolor{MyGold}{rgb}{0.75,0.6,0.12}
\definecolor{MyDarkgray}{rgb}{0.66, 0.66, 0.66}
\definecolor{JiayuanColor}{rgb}{0.60,0.43,0.48}

\newcommand{\ourmethod}{{\sc bilba}}
\newcommand{\ourmethodlong}{{\it bi-level belief assembly}}
\newcommand{\Fenv}{F^{\textrm{env}}}
\newcommand{\Fmanip}{F^{\textrm{manip}}}
\newcommand{\fbel}{f_{\textrm{bel}}}
\newcommand{\bcurr}{b_{\textrm{curr}}}
\newcommand{\Cobs}{C_{\textrm{obs}}}
\newcommand{\Dbel}{D_{\textrm{bel}}}
\newcommand{\Fdes}{\mathcal{F}_{\textrm{des}}}
\newcommand{\Fgoal}{\mathcal{F}_{\textrm{goal}}}

\begin{document}
\title{Bi-Level Belief Space Search for Compliant Part Mating Under Uncertainty}
%
%
\author{Sahit Chintalapudi \and
Leslie Kaelbling \and
Tomas Lozano-Perez}
\authorrunning{S. Chintalapudi et al.}
%
\institute{Massachusetts Institute of Technology\\
\email{\{sahit,lpk,tlp\}@csail.mit.edu}}
\maketitle              
\begin{abstract}
The problem of mating two parts with low clearance remains difficult for autonomous robots.  We present \ourmethodlong{} (\ourmethod{}), a  model-based planner that computes a sequence of \textit{compliant motions} which can leverage contact with the environment to reduce uncertainty and perform challenging assembly tasks with low clearance. Our approach is based on first deriving candidate contact schedules from the  structure of the configuration space obstacle of the parts and then finding compliant motions that achieve the desired contacts.  We demonstrate that \ourmethod{} can efficiently compute robust plans on multiple simulated tasks as well as a real robot rectangular peg-in-hole insertion task. 

\end{abstract}

\section{Introduction}

Many commonplace manipulation tasks, such as loading a dishwasher, shelving a book, or building furniture, can be seen as assembly of rigid parts. These tasks are challenging for robots due to the presence of uncertainty. Typically, the clearances in these tasks are smaller than the uncertainty introduced by pose estimation, grasping, and control errors.  Robots can carry out these tasks by leveraging the forces generated by contact with the environment through the mechanism of compliant motion; however, finding an effective compliant motion strategy for a given task in the presence of uncertainty remains challenging. Model-based search algorithms have been developed~\cite{sieverling2017interleaving,pall2018contingent,wirnshofer2018robust} but their range of application is limited by assumptions or computational complexity.  There are a number of learning-based approaches~\cite{kalakrishnan2011learning,schoettler2019deep}, but these are typically not suitable for dealing with novel tasks without additional learning.  

In this paper, we propose a new model-based method, \ourmethodlong{} (\ourmethod{}), for finding a compliant motion sequence that achieves a desired contact between two parts. As input, \ourmethod{} requires descriptions of the part geometry, initial bounds on the uncertainty of the relative poses of the parts, and access to a physics-based simulator. The sequence of compliant motions is expressed in terms of Cartesian gripper stiffnesses and target positions for the robot gripper. We address problems that have residual uncertainty that cannot be further reduced with sensing, so rather than finding a conditional plan or policy that maps observations to actions, we seek an open-loop ``conformant'' plan that will achieve the assembly with high probability over the initial uncertainty distribution.


We structure \ourmethod{} as a bilevel search. We first search for a sequence of target contacts between the parts and then search for compliant motion parameters (stiffnesses) that are likely to produce those contacts.  This structure can improve efficiency, since it ensures that we have a plausible abstract plan before searching for detailed control parameters.
This search is in ``belief space,'' as we model the effects of each action on the system's belief distribution over the world state, approximated by a set of particles.  The search terminates when it reaches a belief state in which the assembly has been constructed with high probability.

The contributions of this paper are:
(1) an efficient method for choosing continuous low-level control parameters (target setpoint and stiffness matrices) guided by Gaussian Process Regression; and (2) an overall bi-level search control structure that uses abstract plans to guide the search for fully parameterized long-horizon conformant plans. We demonstrate the effectiveness of this method in two simulated problems, a 3D square peg-in-hole and a more complex sequential 3D puzzle, as well the square peg-in-hole on a real Franka Panda robot.  Our experiments show that compared to the most effective baseline, \ourmethod{} can quickly form successful plans and tolerate larger uncertainties under reasonable computational constraints.
\section{Related Work}

Since the earliest days of programmable robots, it has been known that, by tailoring the robot's behavior in response to forces, tasks could be robustly accomplished in spite of pose and control uncertainty~\cite{whitney1985historical}.  A great deal of research followed defining control models and algorithms for providing programmable {\em compliant} behavior~\cite{salisbury1980active,2016-siciliano}.  The key challenge has been to match the compliance to the task.  Influential early work showed that for the special case of a cylindrical peg and chamfered hole, an arrangement of springs supporting the peg (the {\em remote-center compliance}) could achieve reliable assembly (via a single insertion motion) in the presence of (limited) initial uncertainty~\cite{simunovic1979information,Whitney1982QuasiStatic}.  Subsequent work aimed to find compliance parameters (such as stiffness or admittance matrices) from a task description, for example,
Peshkin~\cite{peshkin1990programmed} derived an optimization-based strategy for generating stiffnesses in the planar case.
It has become clear, however, that for each manipulation task there is a complex relationship between the geometric and mechanical properties of the task and the compliance and motion commands required for reliable execution.  The problem of determining suitable compliances can be framed as a deep reinforcement learning problem~\cite{kalakrishnan2011learning,martin2019variable} in which the parameters are learned from data, but such learning has to be redone for each new task.  Recent work on policy code generation using large language models~\cite{burns2024genchip} is promising, but it is not yet clear how far it generalizes.

The seminal work of Mason~\cite{mason1981compliance} provided a framework for understanding compliant motion for tasks as specifying the behavior with respect to contact surfaces in the  configuration space of the robot and task.  This led to one of the earliest approaches to planning sequences of compliant motions from descriptions of the task geometry~\cite{lozano1984automatic}, by constructing a sequence of {\em pre-images} of goal regions under compliant motions.  This {\em pre-image backchaining} approach led to follow-on work~\cite{donald1987error,erdmann1986using,QiaoArise} but it has not scaled to realistic settings.  However, the idea of planning compliance by characterizing contact surfaces (modes) and moving between them by specifically designed compliant motions continues to be fruitful.

Dakin and Popplestone~\cite{dakin1992simplified} provided an early example of generating sequences of compliant motions based on a high-level plan of contacts in a simplified model.  Xiao~\cite{xiaocontactspace} generates a complete graph of {\em contact formations} (CF) possible between two polyhedral bodies, where two contact formations are connected by an edge if there exists a compliant motion that transitions the set of contacts from one CF to the other. Meesen et al.~\cite{MeessenCompliantMotion} show how to implement the compliant motions between contact formations when the configuration of the parts is fully observed.  As opposed to building and searching the full CF-graph, some approaches run a forward RRT-like search in configuration space~\cite{cheng2022contact,pang2022global}.  When a new configuration is explored, the set of possible contacts from that configuration is dynamically computed. These strategies have the advantage of differentiating between CFs where particular subsets of the active contacts are sticking or sliding depending on the friction.   However, these strategies do not explicitly model the effect on uncertainty, whether in the initial state or the actuation. While, they inherit some robustness from the compliant motions they do not deal with, for example, substantial initial pose uncertainty which can change the initial CF.

To deal with greater uncertainty, parts-mating has been studied as an instance of more general approaches to reasoning under uncertainty, in particular via formulation as partially observed Markov decision processes (POMDPs).  The key idea is to plan not in the space of robot configurations but in the space of {\em beliefs}~\cite{kaelbling1998planning,hauser2010randomized,platt2017efficient}.  A belief is a representation of the states that the robot and environment may be in, generally in the form of a probability distribution or a set of sample states.  The belief-space planning problem is that of finding a trajectory from an initial belief state to a belief state that satisfies a goal condition. \textit{Conformant plans} do not depend on observations during execution and are a sequence of actions that reach a goal belief state from the initial belief state with high probability. \textit{Contingent plans}, on the other hand, are plans which account for observations by formulating a strategy which branches depending on observations obtained while executing the actions.

In general, finding contingent plans (policies) in continuous state and action spaces, whether by solving POMDPs or by direct policy search, is extremely challenging.  There is relatively little work on finding contingent compliant-motion plans.  One such approach is  ConCERRT~\cite{pall2018contingent}, an RRT-like planning algorithm for grasping and pushing domains in the presence of initial pose uncertainty and noisy actuation.  The approach leverages error-free contact sensing to keep the belief-space search manageable.  Grafflin et al.~\cite{PhillipsGrafflin2017PlanningAR} also use an RRT in belief space to build a partial policy that models the effect of actuation error, but it also assumes the resulting contact state is fully observed.

There are a number of approaches to synthesizing compliant motion sequences that frame the problem as one of conformant planning.  The original pre-image backchaining approach~\cite{lozano1984automatic} produced a conformant plan.  More recent methods, rather than searching backwards from the goal, perform forward search in belief space where the actions are parameterized as compliant motions. Because of the complexity of evaluating robot dynamics and the mechanics of part contact, it is assumed that the planner has access to a simulator to evaluate the results of a compliant motion on a particle (a sampled state) in the belief state. Kim et al.~\cite{KimHeuristic} use a variant of A* search in a discretized version of the belief space. Unlike our method, their strategy requires one set of pre-defined motion primitives both to implicitly generate a contact schedule and another set of engineered motion primitives to search the belief space. Contact-Exploiting RRT~\cite{sieverling2017interleaving} runs an RRT search in belief space, with the requirement that every particle in a belief state has the same set of contacts. In addition to free-space motions, if a belief state's particles are all in contact with the same surface, compliant motions that ensure the robot slides while maintaining contact are also explored.  The work that most resembles our algorithm is the Anytime-Optimal Belief Expansive State Tree planner (AO-B-EST)~\cite{wirnshofer2018robust}. AO-B-EST is a kinodynamic RRT in belief space which does not constrain a belief state to have the same set of contacts for every particle. Unlike AO-B-EST, our work leverages a contact mode schedule to accelerate the forward search and also picks different stiffness parameters depending on the local configuration space geometry.

\section{Background}
\subsection{Assumptions}
We are interested in mating a part held by the robot (the \textit{manipuland}) with a part in the robot workspace (the \textit{environment}). We assume both parts are rigid polyhedra and that their geometry and inertial properties are known. In Section~\ref{sec:geom} we describe the specification of these parts and  the configuration-space obstacles induced by the environment in the configuration space (C-space) of the manipuland. 

We assume that, throughout the assembly operation, the environment is fixed relative to the stationary base of the manipulator and the manipuland is fixed relative to the gripper. While these transformations are fixed, they are not observable to the robot. This motivates the need to be robust over a distribution of possible manipuland and environment poses. Such a formulation gives rise to a search problem in belief space (see Section~\ref{sec:state_space}).

Under these assumptions, we can transcribe the task of robust part mating into a search over stiffness controls that transition an initial set of representative configurations to a set of configurations where the parts are mated (Section~\ref{sec:problem}).

\subsection{Workspace and Configuration-Space Geometry}
\label{sec:geom}
Both the manipuland and the environment can be expressed as the union of closed convex polyhedra. We refer to face $i$ on the manipuland as $\Fmanip_i$ and face $j$ in the environment as $\Fenv_j$. 

The manipuland has three translational degrees of freedom and three rotational degrees of freedom, so its configuration $q_M$ belongs to the Lie group $SE(3)$. We write the translational component of the configuration as $p_M \in \mathbb{R}^3$ and the rotational component as $R_M\in SO(3)$. Due to the rigid body assumption the manipuland can contact, but never penetrate, the environment. The set of configurations that correspond to a non-empty intersection of the manipuland geometry and environment geometry is written as $\Cobs \subset SE(3)$.

\ourmethod{} exploits the structure of the C-space obstacle to guide the search process. Provided the manipuland and environment are both polyhedra, the cross-section of the C-space obstacle at the fixed rotation $R_M$, denoted $\Cobs(R_M) \subset \mathbb{R}^3$, is guaranteed to be a polyhedron. Configurations on the interior of the C-space obstacle, $int(\Cobs)$ are unreachable. However configurations on the boundary of the C-space obstacle, denoted $\partial \Cobs = \Cobs \setminus int(\Cobs)$ correspond to configurations where the manipuland is in contact with the environment. We can annotate the faces of the boundary of a cross-section of the C-space obstacle, $\partial \Cobs(R_M)$ with the different contacts that are active on these faces.

\subsection{The Belief Space}
\label{sec:state_space}
Let $q_r \in \mathbb{R}^7$ define the joint angles of the robot, $q_O\in SE(3)$ define the pose of the environment relative to the robot base, and ${}^Gq_M \in SE(3)$ define the static transform from the gripper to the manipuland. Note that the manipuland configuration $q_M$ can be inferred from $q_r$ and ${}^Gq_M$ via the forward kinematics of the manipulator. Thus the state of the world is fully specified by $q = (q_r, q_O, {}^Gq_M)$. Via the robot's joint sensors, $q_r$ is observed, however we do not have access to the exact values of $q_O$ and ${}^Gq_M$

A configuration is associated with a (possibly empty) set of contacts between the manipuland and the environment. We write the contact mode as a set-valued function $D(q_M) = \{(\Fenv_i, \Fmanip_j), ..., (\Fenv_k, \Fmanip_l)\}$, where $(\Fenv_i, \Fmanip_j)$ denotes contact between face $i$ in the environment and face $j$ on the manipuland. 

A \textit{belief state} is represented as a set of configurations $b = \{q_1, ..., q_N\}$. Each configuration, also referred to as a ``particle,'' corresponds to a possible underlying world state. In this way the belief defines an empirical distribution over the joint C-space of the robot, manipuland, and environment. We denote contacts that are active for every particle in the belief state by defining $\Dbel(b) = \cap_{q \in b} D(q_M)$. Similarly, the set of contacts such that the contact is active for any of the particles in the belief state is given by $\tilde{D}_{\textrm{bel}}(b) = \cup_{q \in b} D(q_M)$.

\subsection{Stiffness Control}
\label{sec:action_space}
A stiffness controller is tasked with sending torques to the robot's joints such that the robot's dynamics emulate that of a spring-damper system~\cite{2016-siciliano}. Such a system is characterized by the stiffness of the spring $K\in\mathbb{R}^6$, the damping coefficient $B \in \mathbb{R}^6$, and the two endpoints of the spring-damper. We set the damping as a function of the spring stiffness; unless otherwise specified, the damping is set as $B = 2\sqrt{K}$. In this work, one endpoint of the virtual spring-damper is the robot gripper, which is uniquely determined by the joint angles $q_r$. The gripper frame is pulled towards the other endpoint of the spring, a fixed spatial frame $G_d \in SE(3)$ referred to as the \textit{setpoint}. We denote a compliant motion as $u = (K, G_d)$ and a \textit{plan} as an ordered sequence of motions $\zeta = (u_1, ..., u_H)$.

A stiffness control defines a closed-loop policy that maps measured joint angles into forces exerted by the robot gripper. Due to sticking friction and possible collision with the environment, this controller may not succeed in driving the gripper frame to the desired pose. Therefore, each compliant motion also has an associated timeout; unless otherwise specified this timeout will be 5 seconds. 

Given a desired gripper pose, we can compute corresponding joint configurations $q_r^d$ via inverse kinematics for a particular particle. While the joint configurations are not uniquely determined by the gripper pose, we can generate an arbitrary configuration that is both consistent with the desired gripper and minimizes a joint centering objective~\cite{manipulation}. Given a Cartesian stiffness $K \in \mathbb{R}^6$, we can derive an equivalent configuration space stiffness matrix parameterized by the current joint angles denoted $K_q$ (see~\cite{paul1980force} for derivation details). Note that while the Cartesian stiffness is fixed, the joint stiffness varies as the robot moves.

Let $\tau_g \in \mathbb{R}^7$ be the vector of torques acting on the robot due to gravity. The closed-loop control law implemented by the controller is given by:
\begin{equation*}
\tau = K_q(q_r^d - q_r) - B_q\dot{q_r} -\tau_g\;\;.
\end{equation*}
This control law is independent of any uncertainty in the state, as it only depends on the fully observed $q_r$. When the planner outputs a gripper pose $G_d$, it also caches a corresponding $q_r^d$ to be used for computing controls during execution. 

\subsection{Evaluating Belief-Space Dynamics}
\label{sec:dynamics}
The dynamics of a configuration under a stiffness control depend both on the parameters of the stiffness and the resistive forces generated by friction and contact.  Predicting the system behavior in response to a command, even for a known configuration, is quite challenging. For this reason, our planner assumes access to a deterministic physics simulator (we use the Drake simulator~\cite{drake}, however the planning algorithm is agnostic to the choice of simulator). We will write the C-space dynamics as $q'= f(q, u)$. Note that because the manipuland is rigidly attached to the gripper and the other part is rigidly fixed in the world, the only component of the state that changes after a motion are the robot joint angles. In other words: $q' = (q_r', {}^Gq_M, q_O) = f((q_r, {}^Gq_M, q_O), u)$.

The belief space dynamics are evaluated by simulating each particle in the belief state under the compliant motion. Thus, for a set of particles $b$ we write:
\begin{equation*}
b' = \fbel(b, u) \Leftrightarrow \{q^{1\prime},..., q^{N\prime}\} = \fbel(\{q^1, ..., q^N\}, u)\;\;.
\end{equation*}

We extend this notation by writing the posterior belief under a plan $\zeta$ as $b' = \fbel(b, \zeta)$.

\subsection{Problem Statement}
\label{sec:problem}
Given the geometries of the manipuland and environment as a union of convex polyhedra, and an initial belief state $b_0 = \{q_1, ..., q_N\}, q_n \in \mathbb{R}^7\times SE(3) \times SE(3)$, our goal is to compute a sequence of stiffness controls $u_1, ..., u_H, u_h \in \mathbb{R}^{6\times 6} \times SE(3)$ such that the resulting belief under these motions satisfies

$$(\Fenv_{d1}, \Fmanip_{d2}) \in \Dbel(\fbel(b_0, (u_1, ..., u_H)))$$
where $(\Fenv_{d1}, \Fmanip_{d2})$ corresponds to a desired contact between face $d1$ in the environment and face $d2$ on the manipuland. Note that any belief that satisfies the goal contact across its particles is considered a successful assembly, but that other contact constraints may also be active. 

\section{Bi-Level Search Algorithm}

Our method, \ourmethod{}, iteratively generates plausible contact sequences that achieve the goal contact and then searches for compliance parameters that satisfy these contact sequences. The pseudocode is provided in Algorithm~\ref{alg:full_alg}.

\begin{figure}
\begin{subfigure}[b]{0.30\textwidth}
\centering
\includegraphics[scale=0.15]{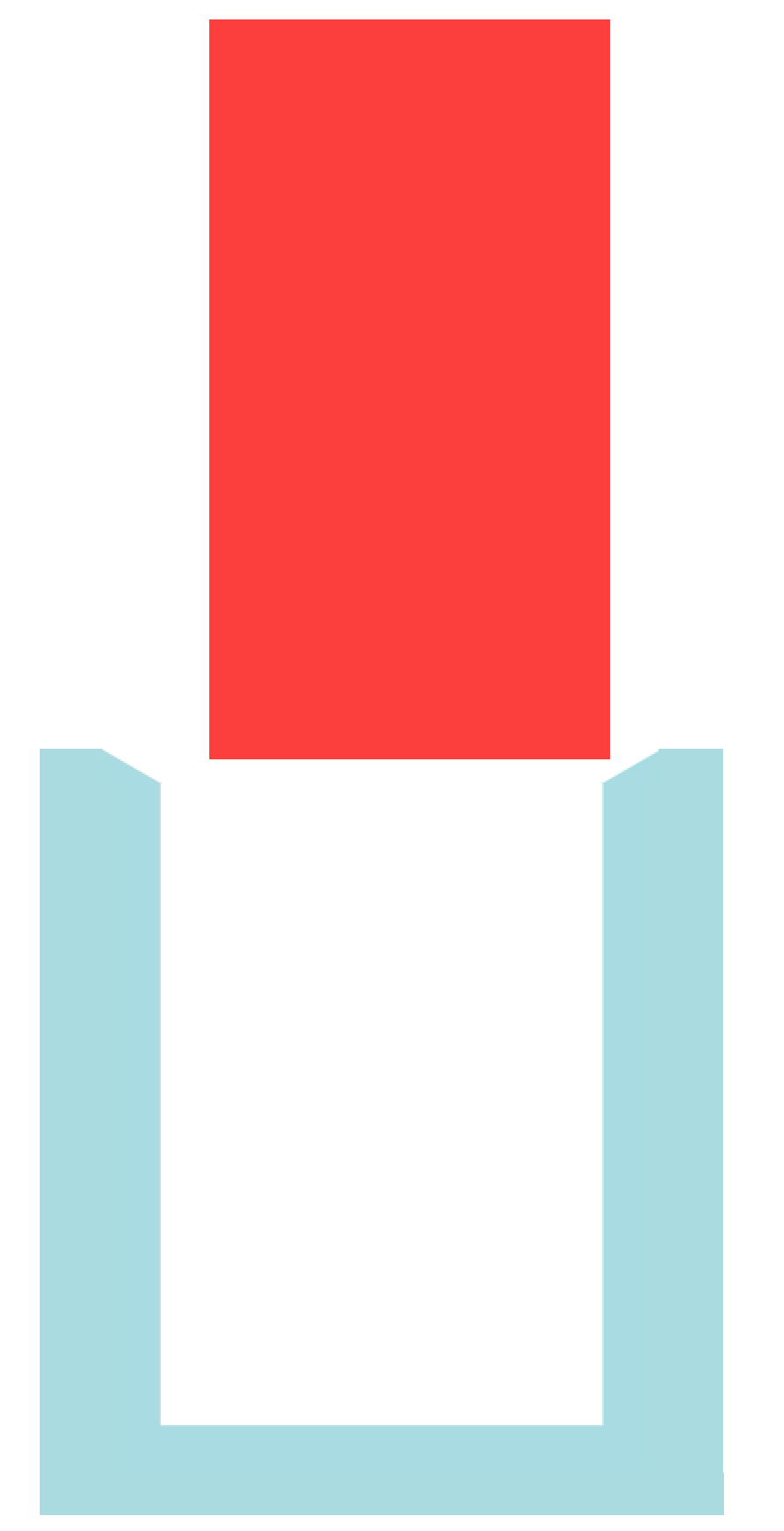}
\caption{Peg and Hole}
\end{subfigure}
\begin{subfigure}[b]{0.36\textwidth}
\centering
\includegraphics[width=\textwidth]{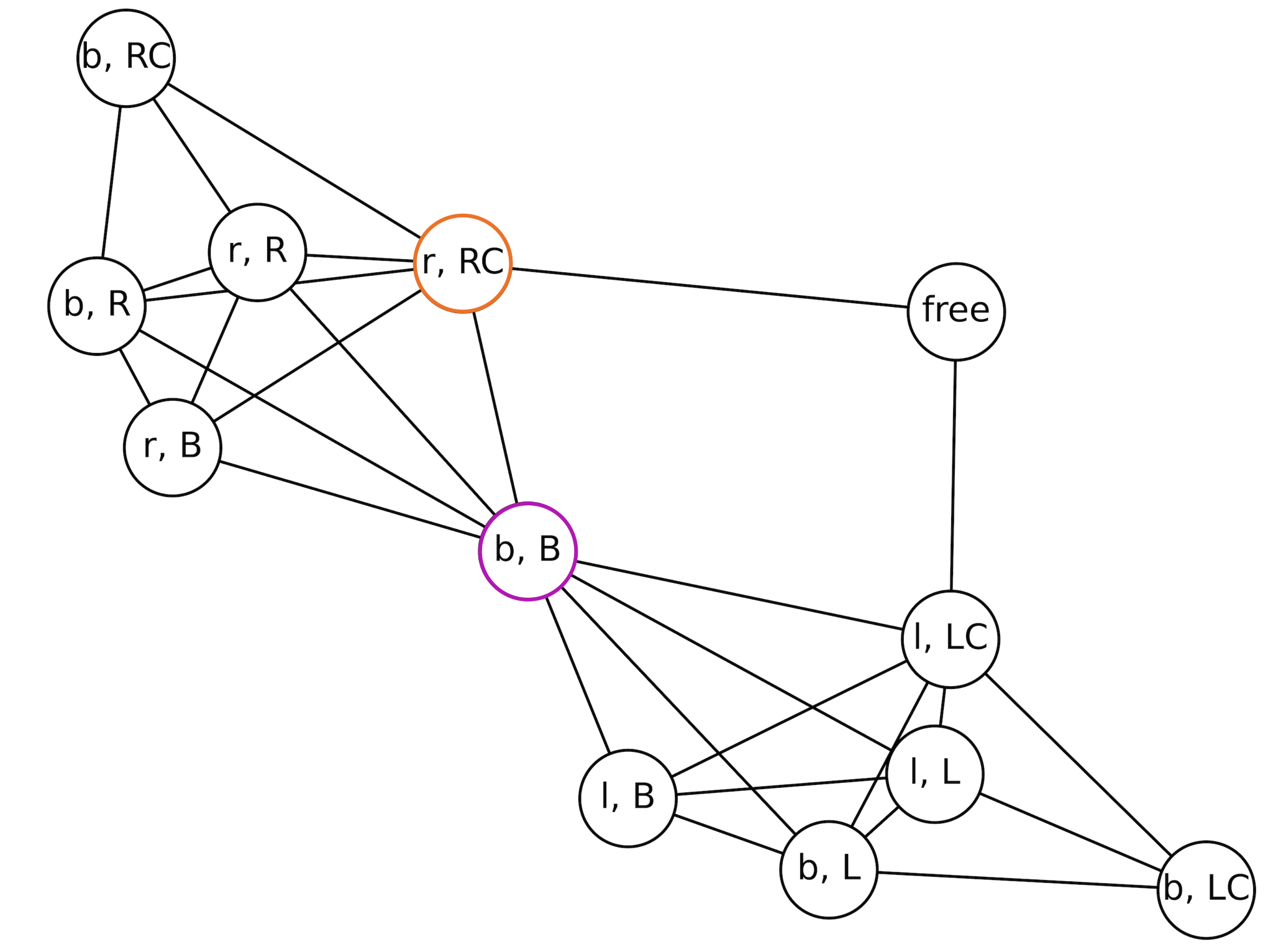}
\caption{Contact Graph}
\label{fig:cg}
\end{subfigure}
\begin{subfigure}[b]{0.30\textwidth}
\centering
\includegraphics[scale=0.17]{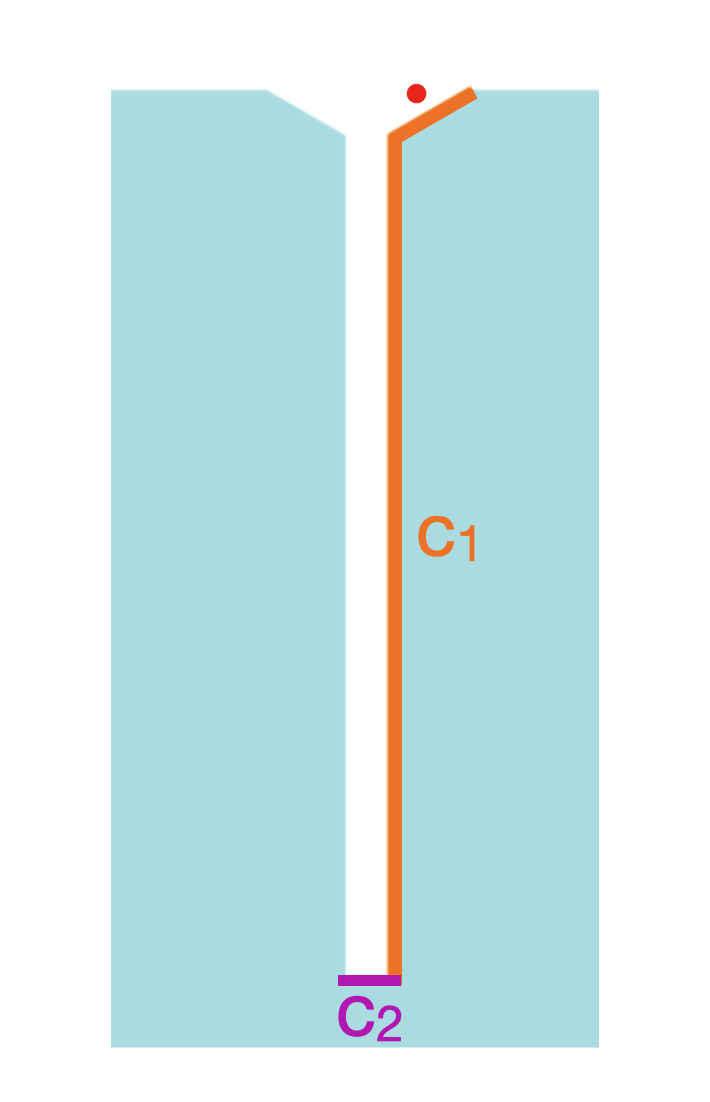}
\caption{Cross section of $C_{obs}$}
\label{fig:cs}
\end{subfigure}

\caption{Let the bottom, left, and right faces of the 2D peg be denoted as "b, l, r" respectively. Similarly, let the faces of the left chamfer, right chamfer, left wall, right wall, and bottom of the hole be denoted "LC, RC, L, R, B". Our method computes a graph over pairwise contacts between faces as shown in~\ref{fig:cg}. In~\ref{fig:cs}, we see how the contact sequence [free, (r, RC), (b, B)], corresponds to a sequence of surfaces in the configuration space which guide the search.}
\label{fig:contact-schedule}
\end{figure}

In Section~\ref{sec:graph}, we construct a discrete graph over contact states (line~\ref{alg:full_alg:make}) which is weighted to prioritize contacts which may reduce uncertainty. This graph is searched to generate candidate contact sequences (line~\ref{alg:full_alg:djikstra}) and comes from the adjacency structure of the surfaces of a cross-section of the C-space obstacle generated by the environment at a fixed orientation corresponding to a sample from the initial belief (see Figure~\ref{fig:contact-schedule}). The candidate contact sequences guide the low-level sampler used to propose compliant motions. 

\begin{figure}
\centering
\begin{subfigure}[b]{0.32\textwidth}
\centering
\includegraphics[width=\textwidth]{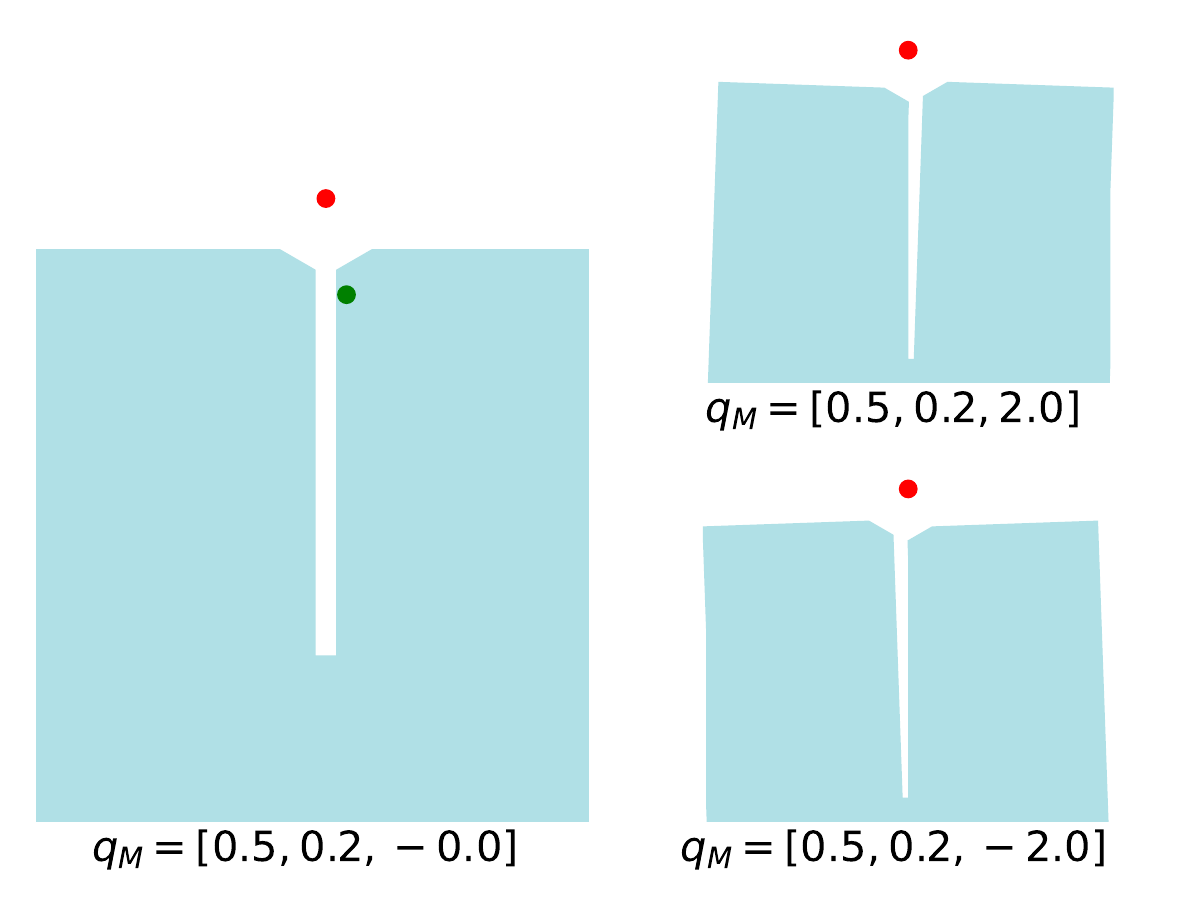}
\caption{}
\end{subfigure}
\begin{subfigure}{0.32\textwidth}
\centering
\includegraphics[width=\textwidth]{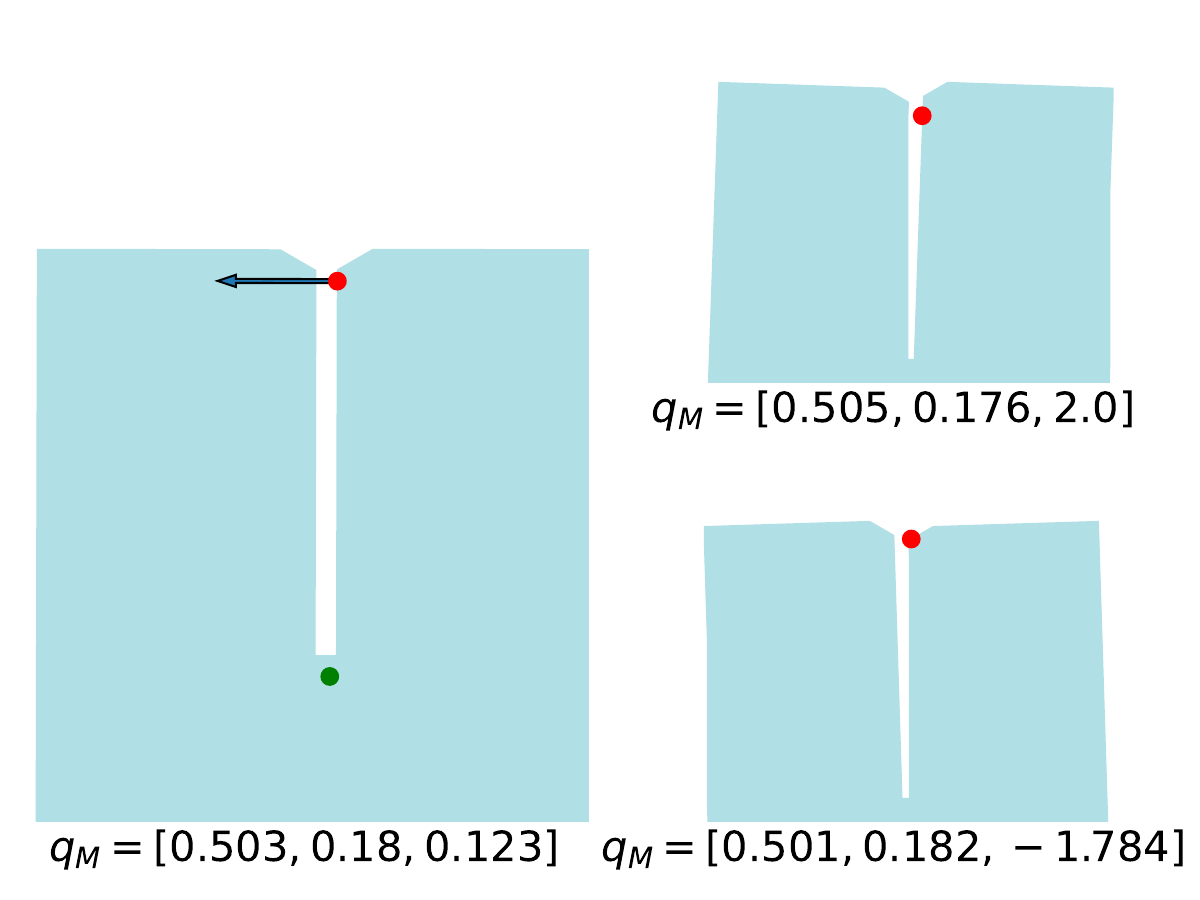}
\caption{}
\label{fig:step_K}
\end{subfigure}
\begin{subfigure}{0.32\textwidth}
\centering
\includegraphics[width=\textwidth]{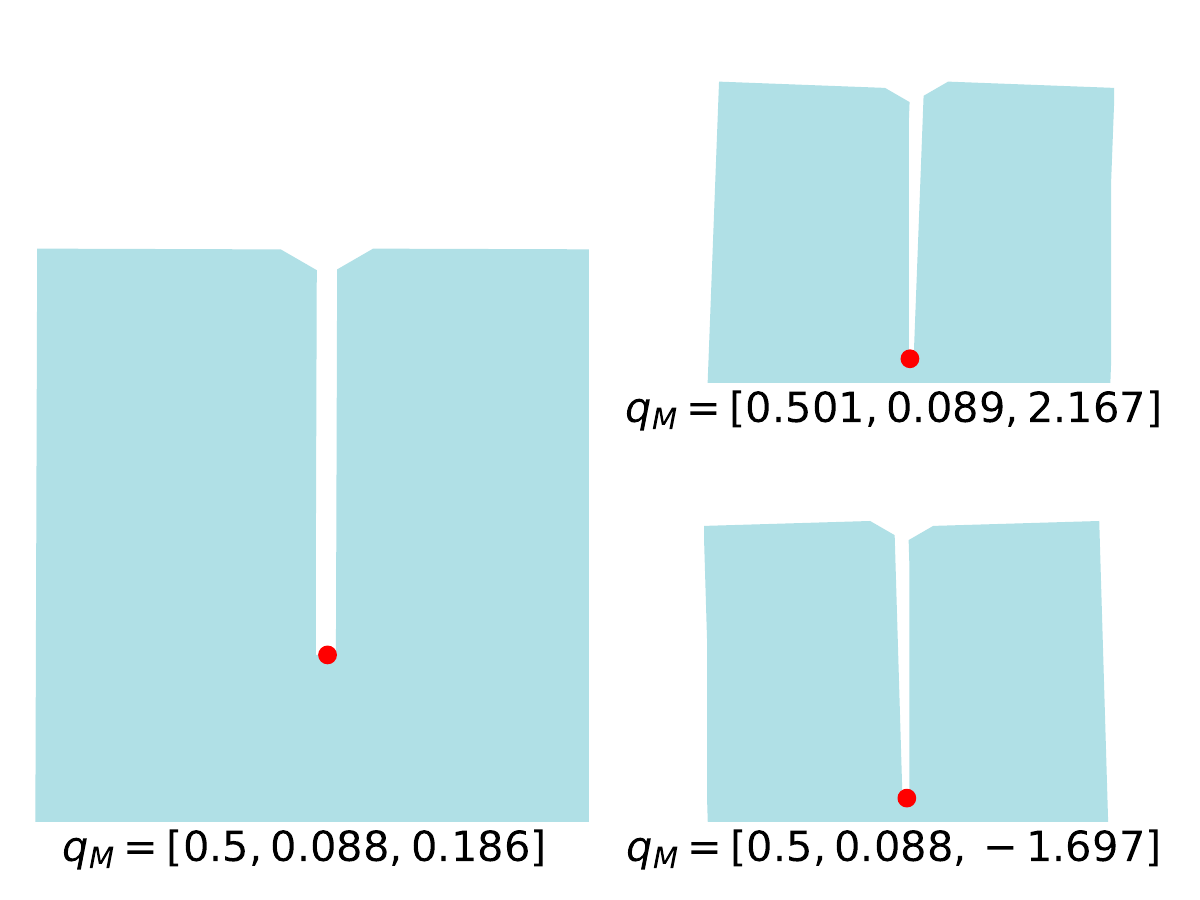}
\caption{}
\end{subfigure}
\caption{A sample 2D belief-space trajectory for peg-in-hole in the presence of rotational uncertainty for three particles. At each execution step, the red points correspond to the $(x, y)$ configuration of the peg for each particle. Because each particle is at a different orientation, the cross-section of $\Cobs$ is slightly different. 
Each green dot corresponds to a setpoint generated by sampling from the surface of $\Cobs$ that is obtained from the abstract contact plan. (a) Initial belief; commanded motion to gain contact with the chamfer.
(b) Belief after first motion;
second motion commanded to bottom of hole, compliant in the x-direction due to the normal forces generated by the surface with which the particle is in contact.
(c) Final belief with all particles in desired state.
}
\label{fig:belief-trajectory}
\end{figure}

Given the subproblem of achieving some intermediate contact from the current belief state, it remains to find a sequence of compliant motions that reaches the contact. The stiffness matrix is generated by calculating the directions normal and tangent to $C_{obs}$ for a particle sampled from the belief. Gripper targets are sampled from the surface of $C_{obs}$ that satisfy the intermediate contact constraint. If the compliance matrix and all sampled gripper targets neither achieve the desired contact nor reduce uncertainty, more candidate gripper targets are sampled from a Gaussian Process conditioned on the already sampled targets, scored on their ability to reduce uncertainty or move subsets of the belief state to the desired contact. This process is detailed in Section~\ref{sec:refine} and corresponds to the \texttt{MakeContact} procedure on line~\ref{alg:full_alg:contact}. We show a 2D representation of the $SE(3)$ motions the sampler produces in Figure~\ref{fig:belief-trajectory}.  Not every edge in the contact graph is guaranteed to admit stiffness controls that transition the entire belief state to the desired contact. If no sequence of compliant motions is found, the edge of the search graph corresponding to the subproblem is pruned (line~\ref{alg:full_alg:prune}). 

Because the edge weights of the graph are parameterized by the current belief (see section~\ref{sec:heuristic}), after a contact is made we recompute edge weights using the posterior belief and update the contact schedule guiding the planner (line~\ref{alg:full_alg:update}).

\begin{figure*}[t!]
\begin{algorithm}[H]
\caption{}
\label{alg:full_alg}
    \begin{algorithmic}[1]
    \Procedure{\ourmethod{}}{$b_0, \Fgoal, \{\Fenv\}, \{\Fmanip\}$}
    \State $(V, E) = $\Call{MakeGraph}{$\{\Fenv\}, \{\Fmanip\}$} \Comment{Make contact graph} \label{alg:full_alg:make}
    \For{$i \in [1, ..., N_{attempts}]$}
        \State $\zeta = [\,]$ \Comment{{Candidate problem solution}}
        \State $b_{curr} = b_0$
        \State $[\Fdes^{(h)}]^H_{h=1} = $\Call{Djikstra}{$V, E, \bcurr, \Fgoal$} \label{alg:full_alg:djikstra}\Comment{Initial contact sequence}
        \For{$\Fdes^{(h)} \in [\Fdes^{(h)}]^H_{h=1}$}
            \State $\zeta_i$ = \Call{MakeContact}{$\bcurr, \Fdes^{(i)}$}\Comment{Candidate subproblem solution} \label{alg:full_alg:contact}
            \If{$\zeta_i \neq \emptyset$} \Comment{{Valid solution found}}
                \State $\zeta$.append($\zeta_i$)\Comment{{Save subproblem solution}}
                \State $\bcurr = \fbel(\bcurr, \zeta_i)$ \Comment{{Compute posterior belief}}
                \State $[\Fdes^{(h)}]^H_{h=h}$ = \Call{Djikstra}{$V, E, \bcurr, \Fgoal$}\Comment{{Update contact seq}} \label{alg:full_alg:update}
            \Else \Comment{{No solution found}}
                \State $E = E \setminus \Fdes^{(i)}$ \Comment{{Prune subproblem graph edge}} \label{alg:full_alg:prune}
                \State \textbf{break}
            \EndIf
        \EndFor
    \If{$\Fgoal \in \Dbel(\bcurr)$}
        \State \Return $\zeta$
    \EndIf
    \EndFor
    
\EndProcedure
\end{algorithmic}
\end{algorithm}
\end{figure*}

\subsection{Constructing a Contact Sequence}
\label{sec:graph}
The high-level search graph is defined by a set of vertices and edges $(V, E)$, with each vertex defined by one face on the manipuland and one in the environment. Each edge is a transition between two contact states and is associated with a belief-dependent edge weight $g_b(u, v)$. Given a graph, cost function, current belief, and goal contact, we can compute a set of desired intermediate contacts which factors the assembly task into subproblems.

\paragraph{Graph construction}
\label{sec:graph_ve}
Given a sample $q_M$ from the initial belief state, we can compute the cross section of the C-space obstacle (see Figure~\ref{fig:contact-schedule}) at the corresponding fixed orientation $R_M \in SO(3)$. We then extract the vertices of the boundary of the C-space obstacle cross section $\mathcal{X} = \{x^{(i)} \in \partial C_{obs}(R_M)\}$. Each vertex acts as a certificate that there exists a non-penetrating configuration $x$ where the manipuland and environment satisfy some set of contacts. We construct the vertex set of the high-level search graph, also referred to as the \textit{mode-subgraph}, $G = (V, E)$ as 
$$V = \{(\Fenv_a, \Fmanip_b) | \exists x^{(i)} \in \mathcal{X}, (\Fenv_a, \Fmanip_b) \in D(x^{(i)})\}\;\;.$$
Each vertex of the mode-subgraph corresponds to a feasible pairwise contact between the manipuland and the environment. Let $u = (\Fenv_a, \Fmanip_b)$ and $v = (\Fenv_c, \Fmanip_d)$; then the edge set of the mode-subgraph is defined as $E = \{(u, v) \in V \times V \mid \exists x^{(i)} \in \mathcal{X}, \{u, v\} \subseteq D(x^{(i)})\}$. An edge connects two modes if there exists a $\Cobs$ vertex which satisfies both modes. 

Because the high-level search graph is computed with respect to the manipuland at a fixed orientation, $G$ defines a subgraph of the true underlying mode-graph. As a consequence, contact transitions that require a rotation of the manipuland are not embedded in $G$, preventing this method from finding solutions with substantial rotation. Furthermore, it is not guaranteed every sample $q_M \sim b$ will generate a graph where the goal contact exists and is reachable from the initial contact. As long as one particle in the initial belief generates a graph on which there are paths from the start to goal contacts, we can cache the generated graph for use throughout the search process.

\paragraph{Edge Weights}
\label{sec:heuristic}
From a given initial belief $b = \{q_1, ..., q_n\}$ we can write a matrix of manipuland configurations $\mathcal{M} = \begin{pmatrix}\mid & \mid & \dots & \mid \\ p^1_M & p^2_M & \cdots & p^N_M\\ \mid & \mid & \dots & \mid\end{pmatrix} \in \mathbb{R}^{3 \times N}$. The eigenvector $u(b)$ corresponding to the largest eigenvalue of the matrix $\mathcal{M}\mathcal{M}^T$ provides a direction of maximal uncertainty. We prefer inducing contacts for which $n(v)$, the vector normal to the face of $\Cobs$ that corresponds to that contact, is aligned with $u(b)$. If the normal vector is aligned with $u(b)$, the forces exerted by the contact are likely to reduce uncertainty in a way that facilitates achieving subsequent steps in the contact sequence. Based on this, we define the cost of an edge $g_b(u, v) = 1 - \delta_{\textrm{proj}} \cdot u(b) \cdot n(v)$, with $\delta_{\textrm{proj}}$ being a small positive weight. 

\paragraph{Search}
\label{sec:ucs}
We refer to the initial contact $D(q) = \emptyset $ as the\textit{ free mode}. Technically, every mode in $G$ can be connected to the free mode, because each mode contains a configuration on the boundary of $\Cobs$. The consequence of this is that many of the paths from the free mode to the goal mode attempt to achieve the goal by moving directly to the goal contact or  contacts neighboring the goal in the search graph. Such short contact sequences are generally less likely to be achievable across the entire belief state as they require the robot to make gross motions without incurring error from spurious contacts encountered during the motion.

In order to accelerate the search, we only connect the free mode to contact modes that are geometrically close to the initial configuration. This can be done by sampling points from $\partial \Cobs(R_M)$ and assigning those points to their corresponding contact modes. The $N_{\textrm{free}}$ contact modes with samples closest to the configuration sampled from the initial belief are connected to the free mode. Connecting these vertices of the graph improves efficiency as if $N_{\textrm{free}} = |V|$, edges corresponding to short paths directly from free space to the goal contact would be attempted before considering more physically plausible contact sequences.

By running Dijkstra's algorithm on this graph using the cost function $g_b$, we can compute a candidate contact sequence from an initial belief to the goal contact; see Figure~\ref{fig:contact-schedule} for a simplified 2D visualization of the 3D problem.
\subsection{Making Contact with Compliant Motion}
\label{sec:refine}
Given a current belief and desired contact, our goal is to find a sequence of stiffness controls that can be applied to the belief with the result that every particle in the belief satisfies the desired contact constraint. Each compliant motion is defined by a stiffness matrix $K$ and a desired gripper frame target $G_d$.  

\paragraph{Computing Translational Stiffness}
From the current belief state $\bcurr$, we sample a configuration $q_{\textrm{curr}}$ and corresponding manipuland pose $q_M$. We use the sampled configuration to determine the chosen translational compliance. 

We want the controller to be compliant in the direction normal to the contact it is making and stiff in the directions perpendicular to the normal, as these are directions in which motion can be generated. For example, Figure~\ref{fig:step_K} shows a simple 2D case where we would want to compliant in the x-direction as this is normal to the wall. Let $R_C$ denote the 3x3 rotation matrix that rotates the world frame such that the z-axis of the frame points in the same direction as the contact normal at $q_{\textrm{curr}}$. We write the translational stiffness matrix as 
$$K_t = R_C \cdot \begin{bmatrix}K_{t, \textrm{stiff}} & 0 & 0\\0 & K_{t, \textrm{stiff}} & 0\\0 & 0 & K_{t, \textrm{soft}}\end{bmatrix} \cdot (R_C)^{-1}\;\;.$$
This can be understood as first rotating displacements into the frame where the contact normal is pointing along the z-axis, computing the forces to be exerted in this frame ($K_{t, \textrm{soft}} > 0$ is a compliant setting for the robot, which is less than $K_{t, \textrm{stiff}}$), and then re-expressing those forces in the world frame.

\paragraph{The Full Stiffness Matrix}
To simplify synthesizing a stiffness matrix, we treat rotational stiffness and translational stiffness as independent of one another. (For a full treatment of coupling rotational and translational stiffness
see~\cite{2016-siciliano}).
Similar to the translational case, we can compute the six-dimensional vector normal to $C_{obs}$ at the sampled configuration via finite differencing. Taking the components of this normal corresponding to infinitesimal rotations about roll, pitch, and yaw of the manipuland gives a 3 dimensional vector along which the robot should be rotationally compliant. Let $R_{C_r}$ denote the change of basis function for the rotational component of displacement, $K_r$ the rotational component of the stiffness matrix, and $K$ the full stiffness matrix. We can write:

$$K_r = R_{C_r} \cdot \begin{bmatrix}K_{r, \textrm{stiff}} & 0 & 0\\0 & K_{r, \textrm{stiff}} & 0\\0 & 0 & K_{r, \textrm{soft}}\end{bmatrix} \cdot R_{C_r}^{-1}, \quad\quad 
K = \begin{bmatrix}K_r & 0_{3\times 3}\\0_{3\times 3} & K_t & \end{bmatrix} \in \mathbb{R}^6\;\;.$$

\paragraph{Sampling Gripper Targets}
\label{sec:solve_sp}
It is not guaranteed that there exists a single compliant motion that drives every particle in the belief to the desired contact state. We first consider a relaxation of this problem: finding a single compliant motion that maximizes the number of particles that achieve the desired goal contact. We do this by (1) computing stiffness controls that are successful for individual particles, (2) scoring these motions based on how much of the belief they move to the desired contact, and (3) using these scores to bias gripper target search until a contact-satisfying target is found or a fixed number of iterations is exceeded. 

Because the stiffness matrix $K$ is determined based on the configuration of a sample from $\bcurr$, it is independent of whatever the choice of gripper target is. For this reason, steps (2) and (3) can evaluate any candidate target $G_d$ by simulating the corresponding candidate motion $u = (K, G_d)$.

\paragraph{Gripper Targets for Individual Particles}
\label{sec:sampling}
For an individual particle with a given stiffness, computing a gripper target that achieves the desired contact can be framed as first searching for a manipuland pose $q_M$ that satisfies the contact constraint and then commanding the gripper to achieve $G_d = q_M \cdot ({}^Gq_M)^{-1}$. Due to error from sticking contact with the environment and spurious additional collisions with the environment, not every value of $q_M$ that satisfies the contact constraint corresponds to a stiffness control where the contact is actually achieved. This motivates sampling many possible values for $q_M$ from $\partial \Cobs$.

Because we are using a stiffness controller, we can also consider values of $q_M$ that lie inside $\Cobs$. While these manipuland poses correspond to configurations where the manipuland is penetrating the environment, the controller regulates interaction forces between the manipuland and the environment. The behavior resulting from sampling invalid manipuland configurations is useful as it causes the environment to exert forces on the manipuland that aid in reducing uncertainty. For some fraction $\epsilon_{\textrm{noise}}$ of the sampled configurations, we apply a noise vector drawn from a uniform distribution over $se(3)$. If the resulting configuration remains inside $\Cobs$, we keep the noised version of the configuration.

\paragraph{Scoring Motions}
A stiffness control successfully driving a single particle to the desired configuration does not necessarily imply that the same stiffness control will drive every particle in the belief state to the goal configuration. When considering new stiffness controls, we would like to bias our search to be near motions that are more successful across the belief state.

We evaluate this success based on two criteria.
First, we prefer motions that achieve the desired contact among a higher fraction of particles.
Second, for the motions that achieve success among the highest fraction of particles, we would like a motion that minimizes the uncertainty of the belief state. There are many ways to quantify the uncertainty of the belief, but we define it in terms of the homogeneity of contact within a belief. Each particle within a belief has a discrete set of active contacts between the manipuland and environment. We can look at the Intersection-Over-Union (IOU) to measure the similarity of contact sets across different particles. We prefer minimizing this definition of uncertainty relative to contact as opposed to one more grounded in spatial uncertainty (i.e., directly computing the variance of the poses) because it is the presence of differing contacts that leads to different configuration-space dynamics from different initial states. This gives the following expression for computing the score of a motion applied to a belief relative to a contact pair $\mathcal{F}_{des} = (F^{obj}_i, F^{manip}_j)$:
\begin{equation}
h(b, \Fdes) = \bigg[\sum_{q_w \in b} I(\Fdes \in D(q_w))\bigg] + \lambda \cdot IOU(\Dbel(b), \tilde{D}_{\textrm{bel}}(b))\;\;.
\label{eqn:score}
\end{equation}
With $\lambda \in (0, 1)$ and $I$ being and indicator function which takes value 1 if the contact is achieved for that particle and zero otherwise.

\paragraph{Sampling via Gaussian Process Regression}
If any of the motions generated by the sampling described in Section~\ref{sec:sampling} achieve the current desired contact for every particle in the belief state, the planner can  begin searching for a motion to achieve the next desired contact in the contact sequence. When none of the candidate motions are successful across the belief, we want to use the scores of the previously sampled candidates to bias sampling towards motions with higher scores. We do this by learning a Gaussian Process (GP) that estimates $h(b, \cdot)$ and can be evaluated more cheaply than a call to the simulator. Using this model, we can limit simulation calls only to motions that we predict are likely to maximize the number of particles that achieve the desired contact.


A GP is defined in terms of a mean function $m: \mathbb{R}^n \rightarrow \mathbb{R}$ and kernel function $k: \mathbb{R}^n \times \mathbb{R}^n \rightarrow \mathbb{R}$. Score data is normalized to have zero mean and unit variance, so we fix $m(\cdot) = 0$. For the kernel, we choose the radial basis function $k(x_1, x_2) = \sigma^2\exp(-\ell||x_1 - x_2||^2)$. The values of $\sigma, \ell$ are chosen to maximize the log-marginal-likelihood of the data. We define $\mathcal{K}: \mathbb{R}^{n\times m} \rightarrow \mathbb{R}^{n\times n}$ as the kernel Gram matrix with $\mathcal{K}(X_1, X_2)_{ij} = k(X_{1i}, X_{2j})$. We represent each of the sampled motions as feature vectors via the logarithmic map from $G_d \in SE(3)$ to the set $\mathcal{U} \subset \mathbb{R}^6$. We write the normalized vector of the scores $h(\cdot)$ as $\mathcal{H}$. 

We generate new sample controls to test by applying random noise to samples from $\mathcal{U}$. We generate $N_{GP}$ random controls which will be denoted $\mathcal{T}$. The estimated value of the normalized scores of the controls in $\mathcal{T}$ is given by $\mathcal{H}_{GP} = (\mathcal{K}(\mathcal{U}, \mathcal{U}^{-1}) \cdot \mathcal{K}(\mathcal{U}, \mathcal{T}))^T \mathcal{H}$~\cite{rasmussen2006gaussian}.

We then take the $N_{sim}$ best scored motions and evaluate $\fbel$ over these motions from the current belief. If any of the motions are successful in achieving the desired contact we can move to the next step in the contact schedule. Otherwise, each of the $N_{sim}$ motions result in a posterior belief that can be scored via equation~\ref{eqn:score}. The additional data of motions and scores can be used to retrain the GP to evaluate a new batch of sampled motions. This process is repeated for a fixed number of rounds or until a successful motion is found.

\paragraph{Iterative Improvement}
\label{sec:repeat}

A single stiffness control may not be sufficient for driving the belief to the desired contact configuration. For this reason, we iteratively search for a motion that maximizes the number of particles that satisfies the contact in the posterior belief and then start a new round of search from this posterior. We decide to terminate the search for a compliant motion if none of the candidate motions generated via a fixed number of rounds of Gaussian Process Regression (GPR) increase the number of particles in the the posterior that satisfy the desired contact. Appendix 1 provides pseudocode elaborating on how we interleave sampling from $\Cobs$ and sampling via GPR to greedily increase the posterior score under a sequence of compliant motions.

\section{Experiments}
\label{sec:experiments}
\begin{figure}
\centering
\begin{subfigure}[b]{0.24\textwidth}
\centering
\includegraphics[scale=0.15]{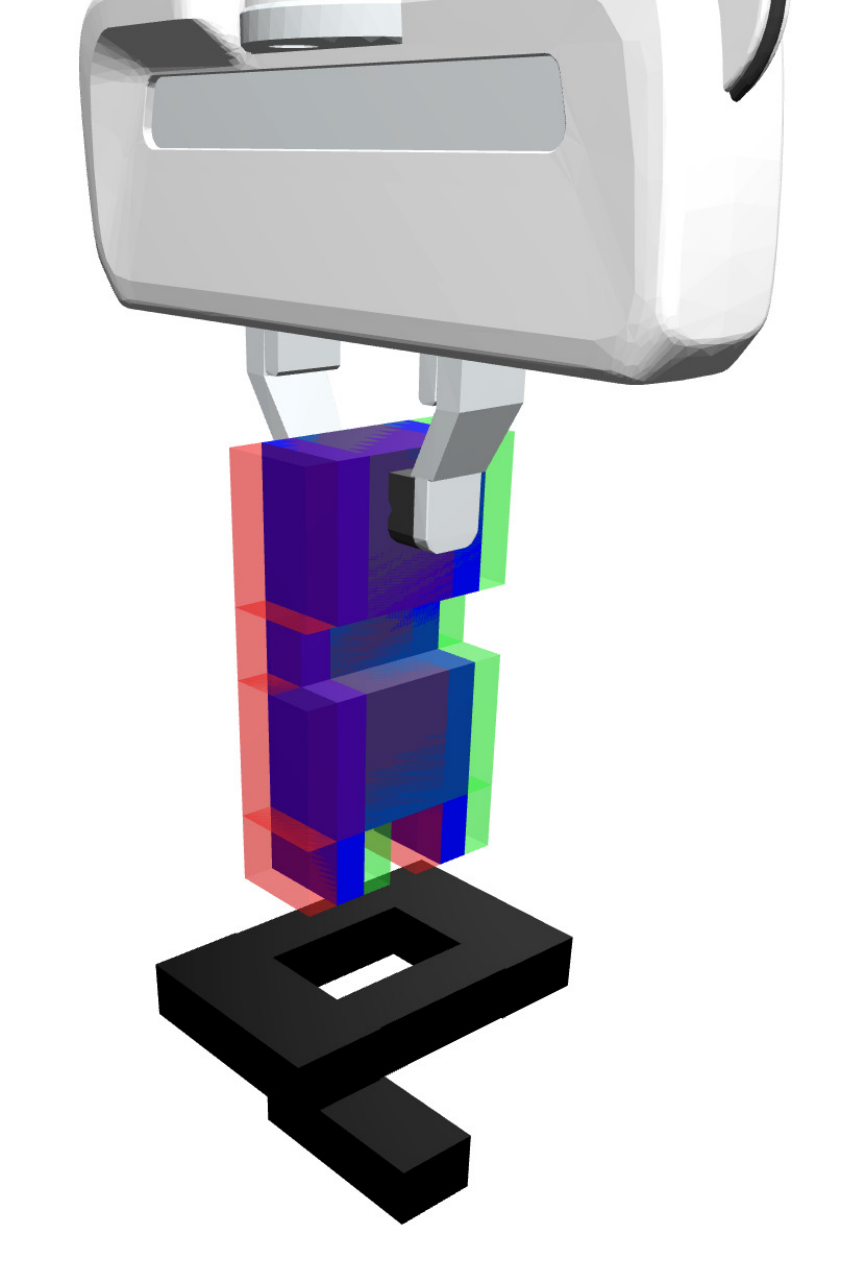}
\caption{Puzzle Start}
\end{subfigure}
\begin{subfigure}{0.24\textwidth}
\centering
\includegraphics[scale=0.15]{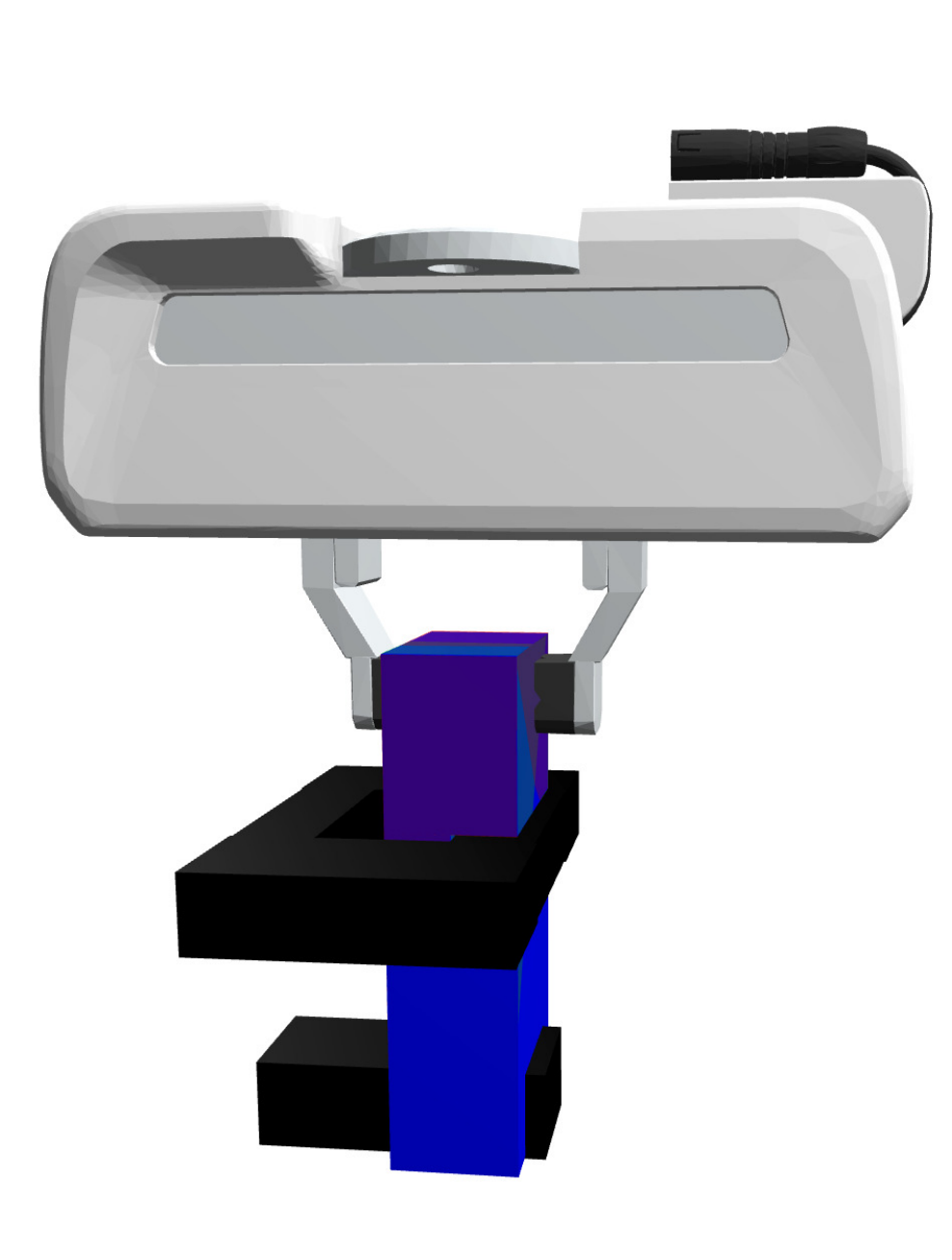}
\caption{Puzzle Goal}
\end{subfigure}
\begin{subfigure}{0.24\textwidth}
\centering
\includegraphics[scale=0.15]{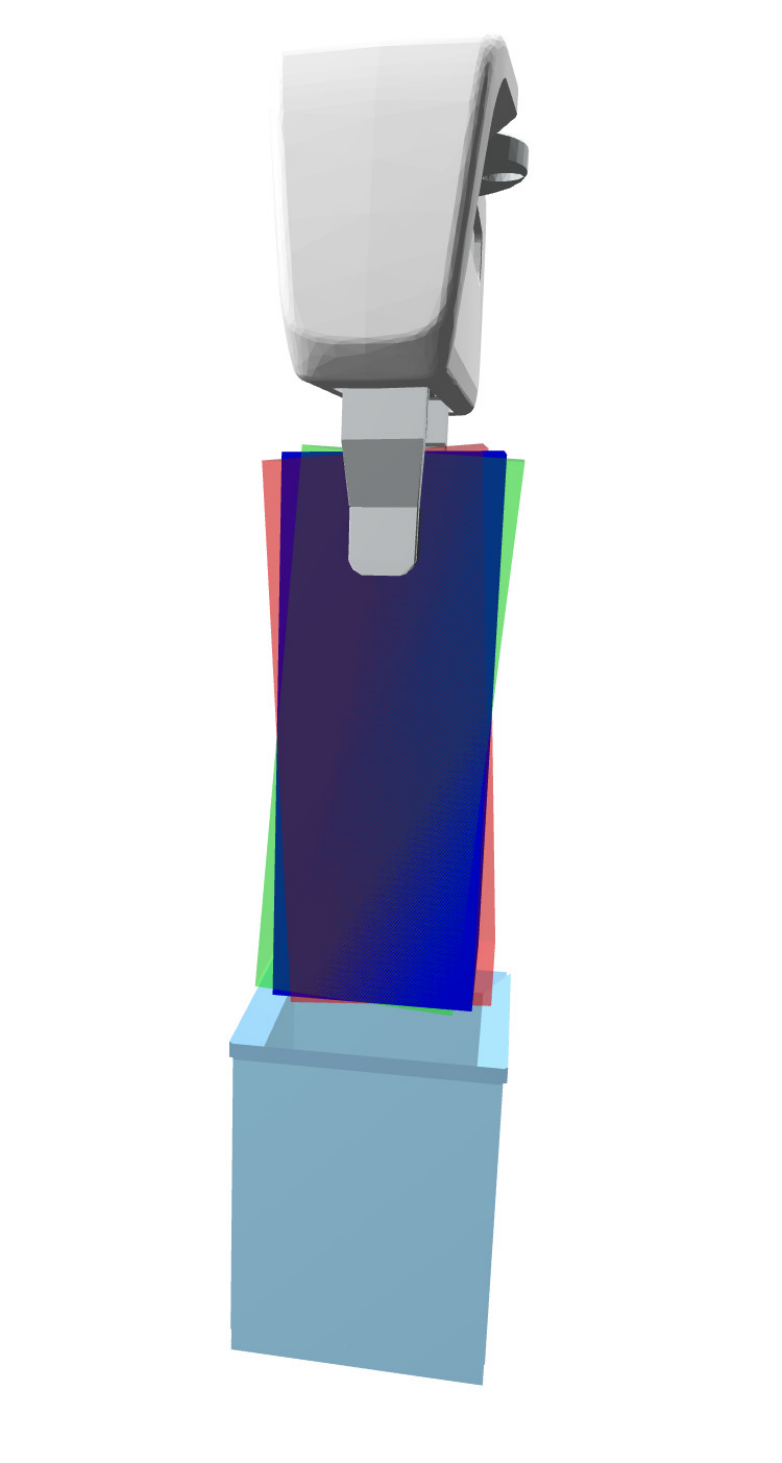}
\caption{Peg Start}
\end{subfigure}
\begin{subfigure}{0.24\textwidth}
\centering
\includegraphics[scale=0.15]{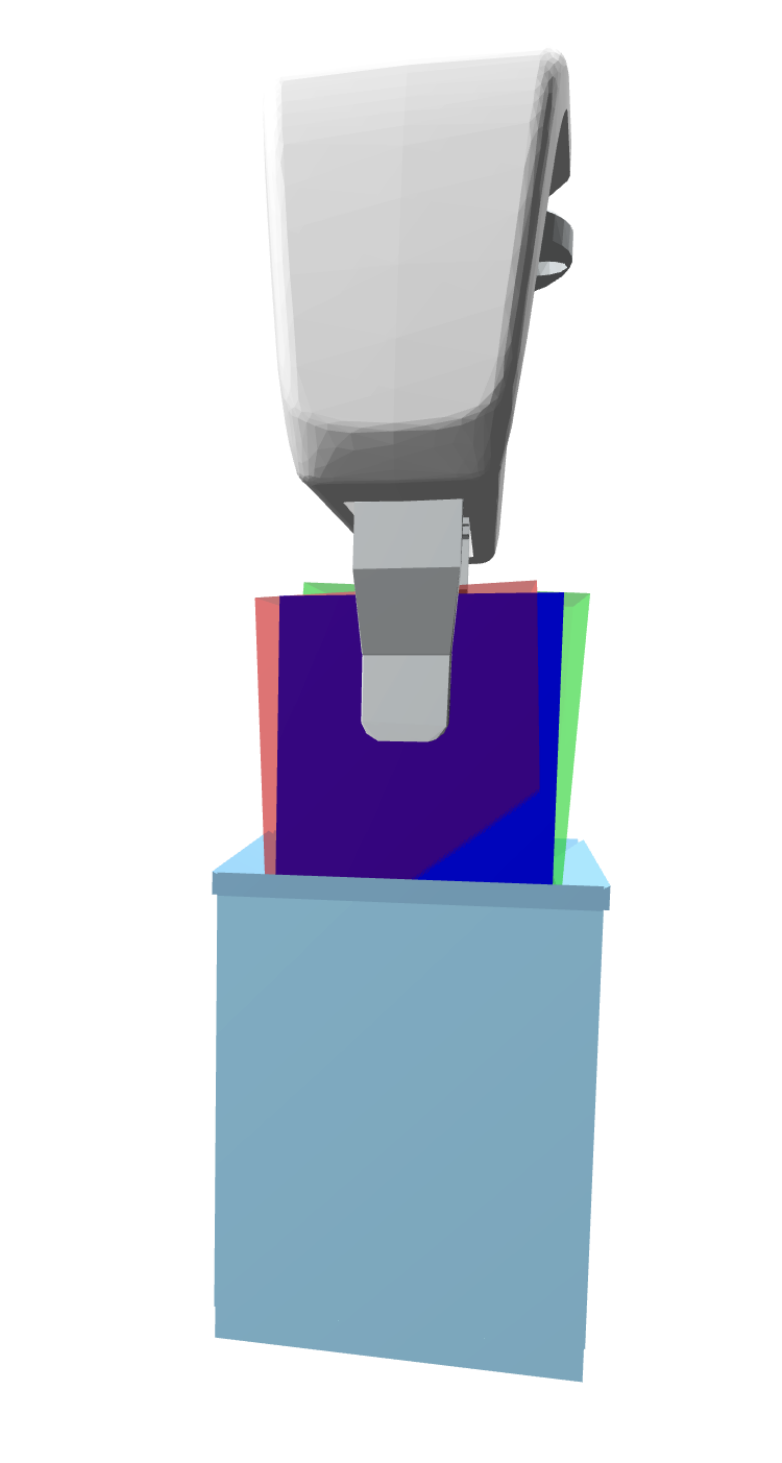}
\caption{Peg Goal}
\end{subfigure}
\caption{Figures 3a and 3b show the ``puzzle'' problem in the presence of translational grasp uncertainty in the x-direction. The red, green, and blue manipulands correspond to three particles representing possible manipuland configurations. Similarly, figures 3c and 3d show the ``peg'' problem in the presence of rotational grasp uncertainty; the possible pitch of the peg is represented by three particles. } \label{fig_exps}
\end{figure}

We validate our method in simulation on two insertion tasks, ``peg-in-hole'' and ``puzzle.'' Peg-in-hole requires inserting a rectangular peg into a chamfered rectangular hole with 5mm of clearance. The puzzle, (proposed in~\cite{dakin1992simplified}) requires vertical alignment with a bottom railing before the manipuland can be mated horizontally with the desired face. Both problems are simulated in Drake (Figure~\ref{fig_exps}). 

We consider two variants of our method. \ourmethod{-Fixed} uses a default set of joint stiffnesses invariant to the current configuration.  \ourmethod{-Variable} computes joint stiffnesses based on the local geometry of the configuration space. We also consider the  Belief-EST (B-EST)~\cite{wirnshofer2018robust} planner, a kinodynamic belief-space RRT that seeks to cover the entire belief space to find a conformant plan from the start state to the goal state without factoring the search into a contact sequence.

We define the initial belief state in terms of 3 particles that bound our uncertainty. For one particle, the parts are rotationally aligned and the manipuland is centered inside the robot's hand. For pitch uncertainty ($\pm n^\circ$) and $x$ uncertainty ($\pm n$ cm), we instantiate one particle with the grasp offset relative to the robot hand by $+n$ and one particle offset relative by $-n$. For y-direction uncertainty by ($\pm n$ cm), the environment is translated parallel to the robot's fingers. Experiments involving more particles where multiple directions of uncertainty are simultaneously active are conduct in Appendix 2.

In the following table we report the median planning time in seconds and median absolute deviation for our two methods and B-EST on the peg-in-hole task. We run each planner ten times from the initial belief state.

 \vspace{10px}

{\centering

\begin{tabular}{|l|l|l|l|}
\hline
& \texttt{\ourmethod{-Variable}} & \texttt{\ourmethod{-Fixed}} & B-EST\\
\hline
Pitch $\pm 1^\circ$ & $\mathbf{10.478 \pm 0.442}$ & $\mathbf{10.247 \pm 0.611}$ & $133.955 \pm 39.587$\\
Pitch $\pm 2^\circ$ & $\mathbf{11.046 \pm 0.859}$ & $\mathbf{10.925 \pm 1.197}$ & $118.031 \pm 35.796$\\
Pitch $\pm 3^\circ$ & $\mathbf{15.365 \pm 2.353}$ & $\mathbf{21.149 \pm 1.711}$ & $264.086 \pm 88.341$\\
Pitch $\pm 4^\circ$ & $70.035 \pm 39.581$ & $\mathbf{33.058 \pm 9.235}$ & $326.984 \pm 106.6266$ \\
\hline
$x \pm 1$cm & $\mathbf{20.125 \pm 2.468}$ & $\mathbf{21.054 \pm 1.063}$ & $186.954 \pm 42.801$\\
$x \pm 2$cm & $\mathbf{23.362 \pm 0.897}$ & $39.625 \pm 13.687$ & $324.610 \pm 125.836$\\
\hline
$y \pm 1$cm & $\mathbf{29.306 \pm 4.749}$ & $52.163 \pm 13.646$ & $154.711 \pm 110.286$\\
$y \pm 2$cm & $\mathbf{101.856 \pm 43.911}$ & $\mathbf{63.38 \pm 15.897}$ & $383.281 \pm 132.070$\\
\hline
\end{tabular}

}

\vspace{6px}

\ourmethod{-Variable} and \ourmethod{-Fixed} exhibit generally similar performance across a wide range of initial uncertainties. For two settings of translational uncertainty ($x \pm 2$cm, $y \pm 1$cm), \ourmethod{-Variable} generates plans faster than \ourmethod{-Fixed}. This suggests that for these settings geometry-dependent compliances enable useful interactions with the surfaces of the hole. For larger uncertainty (pitch $\pm 4^\circ$), \ourmethod{-Fixed} outperformed \ourmethod{-Variable}. In the rotational case, the peg frequently is in multiple-point contact, which means that the configurations of the peg may lie on an edge (intersection of faces) of $C_{obs}$ rather than a single face. In these cases there is not a unique normal vector and the strategy used by \ourmethod{-Variable} may not be appropriate for computing stiffness.

Runtimes for the puzzle task are reported below, * represents timeout.

\vspace{6px}

{\centering

\begin{tabular}{|l|l|l|l|}
\hline
& \texttt{\ourmethod{-Variable}} & \texttt{\ourmethod{-Fixed}} & B-EST\\
\hline
Pitch $\pm 1^\circ$ & $\mathbf{155.558 \pm 92.121}$ & $\mathbf{117.745 \pm 55.822}$ & $877.72 \pm 22.8$\\
Pitch $\pm 2^\circ$ & $\mathbf{455.588 \pm 200.014}$ & $\mathbf{741.13 \pm 194.504}$ & *\\
\hline
$x \pm 1$ cm & $\mathbf{260.314 \pm 131.025}$ & $\mathbf{322.438 \pm 136.585}$ & *\\
\hline
$y \pm 1$ cm & $\mathbf{566.448 \pm 365.694}$ & $\mathbf{543.427 \pm 241.431}$ & *\\
\hline  
\end{tabular}   

}

\vspace{6px}

Across all of our experiments \ourmethod{}-Variable and \ourmethod{}-Fixed outperform B-EST, which illustrates the utility of forming abstract plans to guide the search. For the puzzle task, which requires more steps for assembly, coverage of the entire belief space (as is the strategy of RRT type planners) is intractable when the uncertainty is large. This task was particularly challenging because the parts were unchamfered, so only a small range of forces would be able to mate the parts together. \ourmethod{} directly samples target configurations from surfaces that assist in assembly, leading to solutions found within a tractable runtime. 

\begin{figure}
\centering
\begin{subfigure}[b]{0.49\textwidth}
\centering
\includegraphics[width=0.50\textwidth]{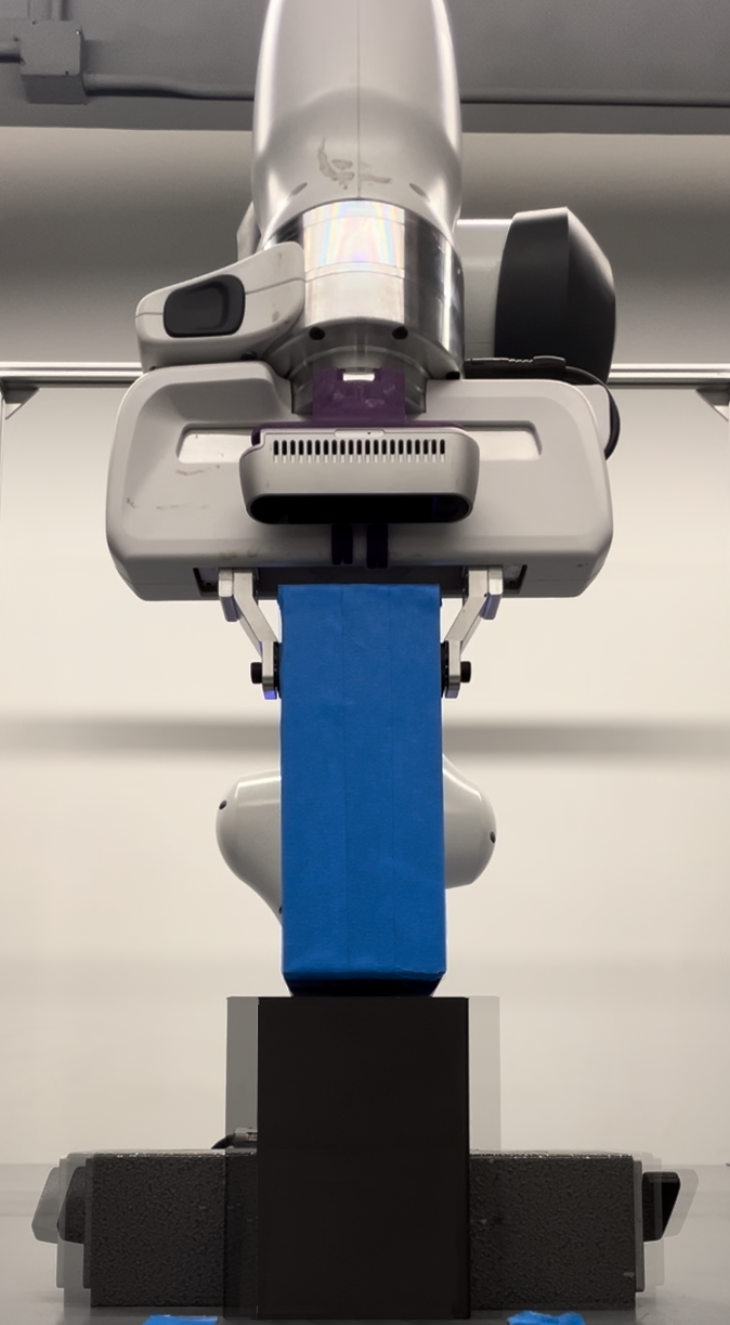}
\caption{The initial robot configuration}
\label{fig:real_init}
\end{subfigure}
\begin{subfigure}{0.49\textwidth}
\centering
\includegraphics[width=0.50\textwidth]{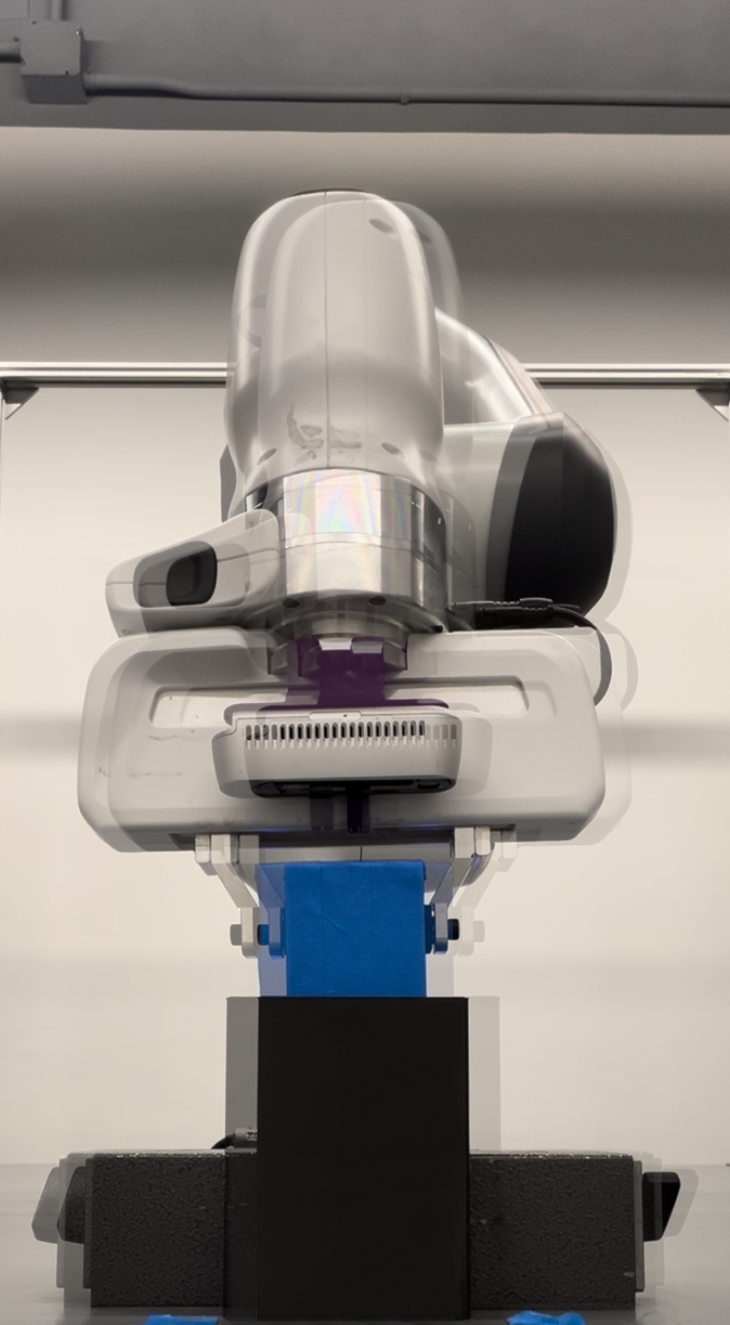}
\caption{Goal robot configuration}
\label{fig:real_goal}
\end{subfigure}
\caption{In figure~\ref{fig:real_init} the manipuland is held at a known configuration relative to the robot but the object position is only known to within 2cm of tolerance (as shown by the two overlaid possible object configurations). After executing the same sequence of compliant motions, the robot is able to successfully insert the peg into the hole in all the initial conditions; as shown in figure~\ref{fig:real_goal}.}
\label{fig:real_experiments}
\end{figure}

\paragraph{Real-Robot Experiments}
We also validate that the trajectory generated by our planner leads to successful execution on a real Franka Panda with a peg rigidly affixed to the gripper. Specifically, we vary the translation of the peg in the y-direction 1 centimeter to the right and 1 centimeter to the left (see Figure~\ref{fig:real_experiments}). 

For the two translational offsets, we ran each of the three planners 10 times, yielding a total of 60 real world insertions. \ourmethod{-Fixed} was successful for 20/20 insertions, B-EST was successful for 19/20 insertions, and \ourmethod{-Variable} was successful for 17/20 insertions. The observed failures were due to compliant motions that induced sliding in simulation actually resulting in a sticking between the peg and hole. One way to address these failures is to use a more conservative estimate of the friction coefficient. Alternatively, we can replan from a new initial belief defined by the current joint angles and original uncertainty bounds. 

\section{Discussion}
We frame robust part mating as a belief space planning problem and propose an efficient strategy for computing compliant motion sequences that are capable of reducing uncertainty and safely making contact with the environment. We validate this approach empirically on real-world and simulated insertions. An open challenge is characterizing the belief-space dynamics of multi-modal, contact-rich systems in a way that enables theoretically guaranteeing the existence of a contact schedule in the mode-subgraph and under what conditions such a schedule admits a sequence of open-loop compliant motions.

To compute cross-sections of the configuration space, \ourmethod{} assumes that the geometries of both the manipuland and environment are fully known. In practice, only the features of the manipuland and environment that will be in contact would be required to generate the relevant surfaces in the configuration space. Framing the assembly problem in terms of only modelling task-relevant parts of each geometry also enables reasoning about shapes that would otherwise require prohibitive computation time. Additionally, the plans \ourmethod{} outputs are an ``open-loop'' composition of closed-loop controllers. Future work could exploit feedback from tactile and force sensors as well as vision to build a contingent policy that depends on the relevant geometric properties of the environment when models are unavailable. Leveraging sensor feedback to build a contingent policy rather than a conformant one may also be useful in domains where the manipuland is prone to slipping in the fingers.

\section{Acknowlegements}
We would like to thank Megha Tippur for assistance with conducting physical robot experiments. We are also indebted to Rachel Holladay for fabricating pieces for assembly and several fruitful conversations.

\bibliographystyle{splncs04}
\bibliography{references}

\begin{thebibliography}{10}
\providecommand{\url}[1]{\texttt{#1}}
\providecommand{\urlprefix}{URL }
\providecommand{\doi}[1]{https://doi.org/#1}

\bibitem{burns2024genchip}
Burns, K., Jain, A., Go, K., Xia, F., Stark, M., Schaal, S., Hausman, K.: Genchip: Generating robot policy code for high-precision and contact-rich manipulation tasks. arXiv preprint arXiv:2404.06645  (2024)

\bibitem{cheng2022contact}
Cheng, X., Huang, E., Hou, Y., Mason, M.T.: Contact mode guided motion planning for quasidynamic dexterous manipulation in 3d. In: ICRA (2022)

\bibitem{dakin1992simplified}
Dakin, G., Popplestone, R.: Simplified fine-motion planning in generalized contact space. In: ISIC (1992)

\bibitem{donald1987error}
Donald, B.R.: Error detection and recovery for robot motion planning with uncertainty. Ph.D. thesis, Massachusetts Institute of Technology (1987)

\bibitem{erdmann1986using}
Erdmann, M.: Using backprojections for fine motion planning with uncertainty. IJRR  \textbf{5}(1),  19--45 (1986)

\bibitem{hauser2010randomized}
Hauser, K.: Randomized belief-space replanning in partially-observable continuous spaces. In: Algorithmic Foundations of Robotics IX, pp. 193--209. Springer (2010)

\bibitem{kaelbling1998planning}
Kaelbling, L.P., Littman, M.L., Cassandra, A.R.: Planning and acting in partially observable stochastic domains. Artificial Intelligence  \textbf{101}(1-2),  99--134 (1998)

\bibitem{kalakrishnan2011learning}
Kalakrishnan, M., Righetti, L., Pastor, P., Schaal, S.: Learning force control policies for compliant manipulation. In: IROS (2011)

\bibitem{KimHeuristic}
Kim, S.K., Likhachev, M.: Parts assembly planning under uncertainty with simulation-aided physical reasoning. In: ICRA (2017)

\bibitem{lozano1984automatic}
Lozano-Perez, T., Mason, M.T., Taylor, R.H.: Automatic synthesis of fine-motion strategies for robots. IJRR  \textbf{3}(1),  3--24 (1984)

\bibitem{martin2019variable}
Martin-Martin, R., Lee, M.A., Gardner, R., Savarese, S., Bohg, J., Garg, A.: Variable impedance control in end-effector space: An action space for reinforcement learning in contact-rich tasks. In: IROS (2019)

\bibitem{mason1981compliance}
Mason, M.T.: Compliance and force control for computer controlled manipulators. IEEE Transactions on Systems, Man, and Cybernetics  \textbf{11}(6),  418--432 (1981)

\bibitem{MeessenCompliantMotion}
Meeussen, W., De~Schutter, J., Bruyninckx, H., Xiao, J., Staffetti, E.: Integration of planning and execution in force controlled compliant motion. In: IROS (2005)

\bibitem{pang2022global}
Pang, T., Suh, H.J.T., Yang, L., Tedrake, R.: Global planning for contact-rich manipulation via local smoothing of quasi-dynamic contact models. TRO  \textbf{39}(6),  4691--4711 (2023)

\bibitem{peshkin1990programmed}
Peshkin, M.A.: Programmed compliance for error corrective assembly. IEEE Transactions on Robotics and Automation  \textbf{6}(4),  473--482 (1990)

\bibitem{PhillipsGrafflin2017PlanningAR}
Phillips-Grafflin, C., Berenson, D.: Planning and resilient execution of policies for manipulation in contact with actuation uncertainty. In: WAFR (2017)

\bibitem{platt2017efficient}
Platt, R., Kaelbling, L., Lozano-Perez, T., Tedrake, R.: Efficient planning in non-gaussian belief spaces and its application to robot grasping. In: ISRR, pp. 253--269. Springer (2017)

\bibitem{pall2018contingent}
Páll, E., Sieverling, A., Brock, O.: Contingent contact-based motion planning. In: IROS (2018)

\bibitem{QiaoArise}
Qiao, H.: {The Combination of Attractive Regions and Pre-images in Motion Planning }. Journal of Manufacturing Science and Engineering  \textbf{124}(2),  341--350 (04 2002)

\bibitem{rasmussen2006gaussian}
Rasmussen, C.E., Williams, C.K.: Gaussian processes for machine learning, vol.~1. Springer (2006)

\bibitem{salisbury1980active}
Salisbury, J.K.: Active stiffness control of a manipulator in cartesian coordinates. In: CDC (1980)

\bibitem{schoettler2019deep}
Schoettler, G., Nair, A., Luo, J., Bahl, S., Aparicio~Ojea, J., Solowjow, E., Levine, S.: Deep reinforcement learning for industrial insertion tasks with visual inputs and natural rewards. In: IROS (2020)

\bibitem{2016-siciliano}
Siciliano, B., Khatib, O. (eds.): Springer Handbook of Robotics. Springer Handbooks, Springer (2016)

\bibitem{sieverling2017interleaving}
Sieverling, A., Eppner, C., Wolff, F., Brock, O.: Interleaving motion in contact and in free space for planning under uncertainty. In: IROS (2017)

\bibitem{simunovic1979information}
Simunovic, S.N.: An information approach to parts mating. Ph.D. thesis, Massachusetts Institute of Technology (1979)

\bibitem{manipulation}
Tedrake, R.: Robotic Manipulation (2023), \url{http://manipulation.mit.edu}

\bibitem{drake}
Tedrake, R., the Drake Development~Team: Drake: Model-based design and verification for robotics (2019), \url{https://drake.mit.edu}

\bibitem{whitney1985historical}
Whitney, D.: Historical perspective and state of the art in robot force control. In: ICRA. ASME (1985)

\bibitem{Whitney1982QuasiStatic}
Whitney, D.E.: Quasi-static assembly of compliantly supported rigid parts. Journal of Dynamic Systems Measurement and Control  \textbf{104},  65--77 (1982)

\bibitem{wirnshofer2018robust}
Wirnshofer, F., Schmitt, P.S., Feiten, W., Wichert, G.v., Burgard, W.: Robust, compliant assembly via optimal belief space planning. In: ICRA (2018)

\bibitem{paul1980force}
Wu, C.h., Paul, R.P.: Manipulator compliance based on joint torque control. In: CDC (1980)

\bibitem{xiaocontactspace}
Xiao, J., Ji, X.: Automatic generation of high-level contact state space. IJRR  \textbf{20}(7),  584--606 (2001)

\end{thebibliography}
\newpage
\section*{Appendix 1: Sampling Score-Maximizing Motions}
As described in Section~\ref{sec:refine}, we compute a sequence of gripper targets that achieves an intermediate contact by greedily trying to maximize a ``score", $h(b, \Fdes)$, that is defined in terms of the number of particles that have achieved the desired contact state and the amount of certainty in the belief. The high-level pseudocode for this procedure is detailed in Algorithm~\ref{alg:makecontact}. 

Each iteration of this optimization involves sampling manipuland configurations on $\partial\Cobs$ that satisfy the desired contact state and then inferring corresponding gripper targets on a particle-by-particle basis. If no samples achieve the desired contact, then we condition a GP on these "training" motions and their corresponding scores. This GP is used to estimate scores for a new batch of "test" motions, the motions that are more likely to achieve high scores are simulated. The pseudocode for both sampling processes is provided in Algorithm~\ref{alg:samplers}.

\begin{figure*}[t!]
\begin{algorithm}[H]
\caption{}
\label{alg:makecontact}
    \begin{algorithmic}[1]
    \Procedure{MakeContact}{$b_{\textrm{curr}}, \mathcal{F}_{\textrm{des}}$}
    \State $K = $\Call{ComputeStiffness}{$\bcurr$}
    \State $h_{\textrm{best}} = 0$
    \State $b = \bcurr$
    \State $\zeta = [\,]$ \Comment{Sequence of score-maximizing motions}
    \While{$\Fdes \not\in \Dbel(b)$}
        \State $u^q, h_{\textrm{cand}}^q, \mathcal{U}, \mathcal{H} = $\Call{SampleMotion}{$\bcurr, K, \Fdes$}\Comment{Try $u$ from $\Cobs$ samples}
        \State $b_{\textrm{cand}}^q = \fbel(b, u^q)$
        \If{$\Fdes \in \Dbel(b_{\textrm{cand}}^q)$} \Comment{Contact achieved by $[\zeta, u^q]$, return}
            \State $\zeta$.append($u^q$)
            \State \Return $\zeta$
        \EndIf
        \State $u^{\textrm{gp}}, h_{\textrm{cand}}^{\textrm{gp}} = $\Call{GPR}{$\bcurr, \mathcal{U}, \mathcal{H},  K, \Fdes$} \Comment{Contact not made, try GP sample}
        \State $b_{\textrm{cand}}^{\textrm{gp}} = \fbel(b, u^{\textrm{gp}})$
        \If{$\Fdes \in \Dbel(b_{\textrm{cand}}^{\textrm{gp}})$} \Comment{Contact achieved by $[\zeta, u^\textrm{gp}]$, return}
            \State $\zeta$.append($u^\textrm{gp}$)
            \State \Return $\zeta$
        \EndIf
        \If {$\max(h_{\textrm{cand}}^{\textrm{gp}}, h_{\textrm{cand}}^q) < h_{\textrm{best}}$} \Comment{Failed to improve score, exit}
            \State \Return $\emptyset$
        \EndIf
        \If{$h^\textrm{gp}_\textrm{cand} > h^\textrm{q}_\textrm{cand}$} \Comment{GPR improved score, continue search from $ b_{\textrm{cand}}^{\textrm{gp}}$}
            \State $h_\textrm{best} = h^\textrm{gp}_\textrm{cand}$
            \State $\zeta$.append($u^\textrm{gp}$)
            \State $b = b_{\textrm{cand}}^{\textrm{gp}} $
        \Else \Comment{$\Cobs$ sample improved score, continue search from $ b_{\textrm{cand}}^q$}
            \State $h_\textrm{best} = h^q_\textrm{cand}$
            \State $\zeta$.append($u^q$)
            \State $b = b_{\textrm{cand}}^q $
        \EndIf
    \EndWhile
\EndProcedure
\end{algorithmic}
\end{algorithm}
\end{figure*}

\begin{figure*}[t!]
\begin{algorithm}[H]
\caption{}
\label{alg:samplers}
\begin{algorithmic}[1]
\Procedure{SampleMotion}{$\bcurr, K, \Fdes$}
\State $h^* = 0$
\State $\mathcal{U} = [\,]$ \Comment{Sampled motion data for GPR}
\State $\mathcal{H} = [\,]$ \Comment{Score data for GPR}
\For{$q \in \bcurr$}
    \For{$n \in N_{\textrm{samples}}$}
        \State $q_M \sim $ \Call{SampleNoisedConfig}{$\Fdes$} \Comment{Manipuland pose satisfiying $\Fdes$}
        \State $G_d = q_M \cdot ({}^Gq_M)^{-1}$ \Comment{Infer gripper target based on particle}
        \State $u = (K, G_d)$
        \State $b' = \fbel(b, u)$
        \State $h' = h(b', \Fdes)$ \Comment{Score $u$ with respect to $\Fdes$}
        \State $\mathcal{U}$.append($u$) \Comment{Update sample data}
        \State $\mathcal{H}$.append($h'$) \Comment{Update score data}
        \If{$h' \geq h^*$} \Comment{Update score-maximizing gripper target}
            \State $h^* = h'$
            \State $u^* = u$
        \EndIf
    \EndFor
\EndFor
\State \Return $u^*, h^*, \mathcal{U}, \mathcal{H}$ 
\EndProcedure

\Procedure{GPR}{$\bcurr, K, \mathcal{U}, \mathcal{H}, \Fdes$}
\State $h^* = 0$
\For{$n \in N_{\textrm{iters}}$}
    \State $\mathcal{T} = $\Call{NoiseMotions}{$\mathcal{U}$, $N_{GP}$} \Comment{Generate $N_{GP}$ test motions}
    \State $\check{U}, \check{T} = $\Call{MatrixLog}{[$\mathcal{U}, \mathcal{T}$]} \Comment{Project train and test data to $\mathbb{R}^6$}
    \State $\check{H} = $\Call{Normalize}{$\mathcal{H}$}
    \State $\mathcal{H}_{GP} = (\mathcal{K}(\check{U}, \check{U^{-1}}) \cdot \mathcal{K}(\check{U}, \check{T}))^T\check{H}$ \Comment{Predict scores of test motions via GPR}
    \State $U_{\textrm{top}} = $\Call{TopN}{$\mathcal{U}, \mathcal{H}_{GP}, N_{sim}$} \Comment{$N_{sim}$ motions with highest predicted scores}
    \For{$u \in U_{top}$}
        \State $b' = \fbel(\bcurr, u)$ \Comment{Simulate high-scoring test motion}
        \State $h' = h(b', \Fdes)$
        \State $\mathcal{U}$.append(u) \Comment{Update dataset with sim results}
        \State $\mathcal{H}$.append(h')
        \If{$h' \geq h^*$} \Comment{Update score-maximizing gripper target}
            \State $u^* = u$
            \State $h^* = h'$
        \EndIf
    \EndFor
\EndFor
\State \Return $u^*, h^*$
\EndProcedure

\end{algorithmic}
\end{algorithm}
\end{figure*}

\section*{Appendix 2: Multi-Dimensional Uncertainty and Compact Belief Representation}
\ourmethod{} takes as input a belief state parameterized as a collection of particles. As the number of particles used increases, computing posterior beliefs under compliant motions requires more simulation time. However, we can show that \ourmethod{} is still capable of efficiently generating fairly robust plans by planning with a simplified representation of the initial belief state.

Consider the peg-in-hole task with uncertainty about every degree of freedom in the grasp. We consider an initial belief state $b_{\textrm{full}}$, where the $x$ position of the peg relative to the gripper is known within 2cm ($\pm 1$cm), the $z$ position is known within 1cm, and the rotation of the peg is known within 6 degrees. We refer to these initial 6 particles as ``extremal" particles. We then randomly sample 8 offsets of the peg relative to the manipuland in terms of x, z, and pitch that lie within the 3-dimensional ellipsoid defined by the x, z, and pitch tolerances.  These randomly sampled offsets define additional particles that, combined with the set of extremal particles, parameterize $b_{\text{full}}$.

Rather than find a plan by passing $b_{\textrm{full}}$ to \ourmethod{}, we can instead construct a more compact belief state $b_{\textrm{plan}}(\alpha)$. This belief state is defined as the union of the set of extremal particles with $\alpha$ samples (drawn without replacement) from the non-extremal particles. $b_\textrm{plan}(0)$ consists only of the extremal particles and $b_\textrm{plan}(8)$ is equivalent to $b_{\textrm{full}}$. We compute 20 plans from $b_{\textrm{plan}}$. For each plan, we then evaluate the number of particles in $b_{\textrm{full}}$ that satisfy the desired contact with the bottom of the hole. The median success rate, as well as the median planning time in seconds, are reported in the table below. 

\vspace{6px}
{\centering

\begin{tabular}{|l|l|l|}
\hline
& Planning Time & Success Rate\\
\hline
$\alpha = 0$ & $35.0 \pm 3.2$ & $0.93$\\
$\alpha = 1$ & $37.8 \pm 3.1$ & $1.0$\\
$\alpha = 2$ & $40.9 \pm 5.4$ & $1.0$\\
$\alpha = 3$ & $44.3 \pm 4.8$ & $0.93$\\
$\alpha = 4$ & $49.1 \pm 3.4$ & $1.0$\\
$\alpha = 5$ & $51.1 \pm 6.7$ & $1.0$\\
$\alpha = 6$ & $82.1 \pm 34.1$ & $1.0$\\
$\alpha = 7$ & $57.1 \pm 10.6$ & $1.0$\\
$\alpha = 8$ & $154.4 \pm 30.7$ & $1.0$\\
\hline
\end{tabular}

}
\vspace{6px}
As expected, we see that as the number of particles that define $b_{\textrm{plan}}$ increase, the planning time also increases. Notably, even without access to $b_{\textrm{full}}$, \ourmethod{} is able to generate plans that at worst still move most of the held-out particles to the desired goal contact. For example, running \ourmethod{} from $b_{\textrm{plan}}(5)$ generates plans that are as robust as running \ourmethod{} from $b_{\textrm{full}}$, but in less than 1/3 the time.

%
%
%

%
%
%
%
\end{document}